\pdfoutput=1

\documentclass[11pt]{article}

\newif\ifarxiv
\arxivtrue 

\ifarxiv
    \usepackage{acl}
\else
    \usepackage[review]{acl}
\fi

\usepackage{times}
\usepackage{latexsym}

\usepackage[T1]{fontenc}

\usepackage[utf8]{inputenc}

\usepackage{microtype}

\usepackage{inconsolata}
\usepackage{enumitem}
\usepackage{array}
\usepackage{amsmath}
\usepackage[capitalise]{cleveref}
\usepackage{listings}
\usepackage{graphicx}
\usepackage{longtable}
\usepackage{booktabs}
\usepackage{soul,xcolor}
\usepackage{longtable}
\sethlcolor{yellow}
\usepackage{subcaption}
\usepackage{multirow}

\definecolor{deepblue}{rgb}{0,0,0.5}
\definecolor{deepred}{rgb}{0.6,0,0}
\definecolor{deepgreen}{rgb}{0,0.5,0}
\definecolor{pink}{RGB}{219, 48, 122}
\definecolor{forestgreen}{RGB}{34,139,34}
\definecolor{goldenrod}{RGB}{218,165,32}
\definecolor{sepia}{RGB}{112,66,20}

\crefname{figure}{Figure}{Figures}
\crefname{table}{Table}{Tables}
\crefname{appendix}{Appendix}{Appendices}
\crefname{section}{Section}{Sections}
\crefname{equation}{Eq.}{Eqs.}

\newcommand\myparagraph[1]{
\vskip 0.05in 
\noindent{\bf {#1}}}
\title{AFaCTA: Assisting the Annotation of Factual Claim Detection\\ with Reliable LLM Annotators}

\author{
    Jingwei Ni\textsuperscript{\rm 1}, 
    Minjing Shi\textsuperscript{\rm 1},
    Dominik Stammbach\textsuperscript{\rm 1},
    Mrinmaya Sachan \textsuperscript{\rm 1}, \\
    \textbf{Elliott Ash}\textsuperscript{\rm 1},
     \textbf{Markus Leippold}\textsuperscript{\rm 2, 3}  \\
    \textsuperscript{\rm 1}ETH Zürich \hspace{5mm}
    \textsuperscript{\rm 2}University of Zürich \hspace{5mm}
    \textsuperscript{\rm 3}Swiss Finance Institute (SFI) \\ 
    \texttt{\{jingni, msachan, ashe\}@ethz.ch, shimin@student.ethz.ch,} \\ \texttt{markus.leippold@bf.uzh.ch}
    }

\begin{document}
\maketitle
\begin{abstract}
With the rise of generative AI, automated fact-checking methods to combat misinformation are becoming more and more important. However, factual claim detection, the first step in a fact-checking pipeline, suffers from two key issues that limit its scalability and generalizability: (1) inconsistency in definitions of the task and what a claim is, and (2) the high cost of manual annotation. To address (1), we review the definitions in related work and propose a unifying definition of factual claims that focuses on verifiability. To address (2), we introduce \textbf{AFaCTA} (\textbf{A}utomatic \textbf{Fa}ctual \textbf{C}laim de\textbf{T}ection \textbf{A}nnotator), a novel framework that assists in the annotation of factual claims with the help of large language models (LLMs). AFaCTA calibrates its annotation confidence with consistency along three predefined reasoning paths. 
Extensive evaluation and experiments in the domain of political speech reveal that AFaCTA can efficiently assist experts in annotating factual claims and training high-quality classifiers, and can work with or without expert supervision. Our analyses also result in PoliClaim, a comprehensive claim detection dataset spanning diverse political topics.\footnote{
\ifarxiv
\url{https://github.com/EdisonNi-hku/AFaCTA}.
\else
We will open-source our code, annotations, and LLM outputs.
\fi
}
\end{abstract}

\section{Introduction}
\begin{table*}[t]
\small
\centering
\resizebox{\textwidth}{!}{
\begin{tabular}{>{\raggedright\arraybackslash}m{0.1\textwidth}m{0.9\textwidth}}
\hline
\textbf{Type} & \multicolumn{1}{c}{\textbf{Examples and Explanations}} \\ \hline

Facts entangled with Opinions & \textit{\textbf{Example 1:}} \hl{We are tackling other needed projects to increase capacity like six-laning I-10 in West Mobile from Theodore to Irvington.} \textit{Fact part: The sentence presents a clear and explicit fact about a project. Opinion part: the project's necessity is a subjective judgment.}

\textit{\textbf{Example 2:}} \hl{We are so thankful that we haven't suffered any loss of life, and it's always heartening to see and hear stories of Alaskans pitching in to help each other.} \textit{Fact part: no people die in the storm (according to contexts).}

\textit{\textbf{Example 3:}} \hl{I thank the legislature for standing with my administration and the people of Alaska by funding this effort.} \textit{Fact part: they fund the effort of resource development (according to contexts).}
 \\ \hline
 
Checkworthy but NOT verifiable & \textit{\textbf{Example 1:}} \hl{Democrats and the Media need to stop using the \#Coronavirus to politicize things and scare people. It's irresponsible. This is not the time to try and gain political points or headlines from scaring people!} \textit{This tweet is labeled as check-worthy by CheckThat!-2021 \citep{nakov2021overview} since it is a polarized political opinion. However, the Democrats' and Media's intention is subjectively interpreted and cannot be verified by objective evidence.}

\textit{\textbf{Example 2:}} \hl{Trump's preference for well-done steaks topped with ketchup.} \textit{This is an unverifiable personal preference. However, it is politicized and used to criticize political figures, thus making it checkworthy.}
\\ \hline

Verifiable but (maybe) NOT check-worthy & \textit{\textbf{Example 1:}} \hl{Italy's Prime Minister Giuseppe Conte has announced that the whole of the country is being put on lockdown in an attempt to contain the \#coronavirus outbreak.} \textit{This tweet with verifiable fact is labeled as NOT checkworthy by CheckThat!-2021.}

\textit{\textbf{Example 2:}} \hl{Zee News: Petrol price reduced by Rs 2.69 CNN: Petrol price reduced by Rs 2.69 BBC: Petrol price reduced by Rs 2.69 NDTV: China is sending Corona Virus to the world via mails and WhatsApp.} \textit{This tweet cites news with verifiable facts. But it is labeled NOT checkworthy by CheckThat!-2021. } 
\\ \hline

Context of Claims & \textit{\textbf{Example:}} ...Those with schizophrenia spectrum and psychosis disorders, many self-medicating with drugs or alcohol addictions.
\hl{That's precisely what our encampment resolution grants and our new CARE Court seek to address.
} Getting people off the streets, out of tents, and into housing and treatment is essential to making our streets safe for everyone, but public safety certainly isn't just about homelessness... \textit{This claim defines the duty of CARE but is not self-contained. It is hard to determine its verifiability without the full semantic information in context.}
\\ \hline




\end{tabular}
}
\caption{\label{tab:examples} Examples that are not well-defined according to definitions in related work, illustrating the definition of factual claim detection is hard and controversial. Example claims are highlighted in \hl{yellow}. Explanations are written in \textit{italics}.}
\end{table*}


The explosion of mis- and disinformation is a growing public concern, with misinformation being widely shared \citep{vosughi_2018}. Manual fact-checking is an important counter-measure to misinformation \citep{lewandowsky2020debunking}. However, fact-checking is a time-consuming and expensive endeavor, and computational remedies are required \citep{vlachos-riedel-2014-fact}.


A first step to identify mis- and disinformation consists of factual claim detection, which filters out the claims with factual assertions that need checking \citep{arslanBenchmarkDatasetCheckWorthy2020,alam-etal-2021-fighting-covid,stammbachEnvironmentalClaimDetection2023}. Considering the sheer amount of daily online content and LLMs' generative capability, we argue that a valid factual claim detection system should be efficient and easily deployable to monitor misinformation consistently. Therefore, we need a way to produce high-quality resources to build transparent, accurate and fair models to automatically detect such claims. However, there are two major challenges in the data collection process.

\myparagraph{Discrepancies in task and claim definitions.} By now, arguably, several different claim definitions exist, which confuse practitioners. 
What is a \textit{claim} is unclear, leading to various \textit{claim detection tasks}, e.g., in automated fact-checking and argument mining. 
For example, \citet{alam-etal-2021-fighting-covid} dismiss all opinions from factual claims, but \citet{guptaLESALinguisticEncapsulation2021} includes ``opinions with social impact'' as factual claims. Many studies \citep{arslanBenchmarkDatasetCheckWorthy2020,nakovOverviewCLEF2022CheckThata} aim at detecting ``check-worthy'' claims while \citet{konstantinovskiyAutomatedFactcheckingDeveloping2020} argues the definition of ``check-worthiness'' is highly subjective and political. Such variances reflect a lack of clarity in conceptualizing critical distinctions, such as the overlap between opinions and verifiable facts (refer to \cref{tab:examples} row 1), and the separate nature of verifiability and check-worthiness in the context of factual claim detection (see \cref{tab:examples} rows 2 and 3). To address these inconsistencies, we propose a definition of factual claims based on verifiability: factual claims present verifiable facts; a fact is verifiable only if it provides enough specificity to guide evidence retrieval and fact-checking. We focus on verifiability to maximize the definition's objectivity and clearly delineate facts from opinions.  

\myparagraph{Manual annotations are expensive.} All existing datasets are manually annotated, which is time-consuming and expensive. Thus, most existing resources are inevitably restricted to certain topics for which it is feasible to annotate claims manually. Such examples include presidential debates \citep{hassan_2015_presidential_debates}, COVID-19 tweets \citep{alam-etal-2021-fighting-covid}, biomedical \citep{wührl2021claim} and environmental claims \cite{stammbach-etal-2023-environmental}. This potentially limits models' ability to generalize to future topics. However, manually annotating datasets with new topics is too expensive. In light of this, we propose \textbf{AFaCTA}, a multi-step reasoning framework that leverages LLMs to assist in claim annotation, making annotation more scalable and generalizable while rigorously following our factual claim definition. 

In fact-checking, it is essential to have high annotation accuracy. However, LLM annotators are far from perfect \citep{ziemsCanLargeLanguage2023,pangakisAutomatedAnnotationGenerative2023}. Thus, to ensure the reliability of LLM annotations, AFaCTA 
calibrates the correctness of the annotations based on the consistency of different paths. 
Our evaluation shows that AFaCTA 
outperforms experts by a large margin when all reasoning paths achieve perfect consistency but fails to achieve expert-level performance on inconsistent samples. 
Nevertheless, we argue that AFaCTA can be an efficient tool in assisting factual claim annotation: perfectly consistent samples can be labeled automatically by the tool, which roughly saves 50\% of expert time (see GPT-4-AFaCTA's perfect consistency rate in \cref{tab:annotation_comparison}). However, inconsistent ones may need expert supervision.

Using AFaCTA, we annotate \textbf{PoliClaim}, a high-quality claim detection dataset covering U.S. political speeches across 25 years, spanning various political topics. We split the 2022 speeches as the test set and the 1998 to 2021 speeches as the training set to imitate the real-world use case where a model learns from the past and predicts future claims. We evaluate hundreds of classifiers trained on various data combinations, finding that AFaCTA's annotated data with perfect consistency can be a strong substitute for data annotated by human experts. 
In summary, our contributions include:

\begin{enumerate}[itemsep=0pt,topsep=1pt]
\item We review the regular misconceptions and confounders in claim definition, proposing a claim definition for fact-checking focusing on verifiability.
\item We propose AFaCTA, an LLM-based framework that assists factual claim annotation and ensures its reliability by calibrating annotation quality with consistency along different reasoning paths.
\item We annotate PoliClaim, a high-quality factual claim detection dataset covering political speeches of 25 years and various topics.
\end{enumerate}

\section{Claim Definition for Fact-checking} \label{sec:definition}
In this section, we first provide an overview of the discrepancies in claim definitions in prior work. Then, we propose our definition of a factual claim with respect to existing discrepancies.

\subsection{Discrepancies in Prior Work} \label{sec:confounders}

\myparagraph{Claim conceptions:} The term ``claim detection'' is used not only in fact-checking but also in other areas of research, for example, argument mining \citep{bolandFactsSurveyConceptualisation2022a}. However, this term refers to different concepts in different research areas. In fact-checking, claim detection aims at identifying objective information in statements, which can be ruled factually wrong or correct according to evidence \citep{thorneFEVERLargescaleDataset2018,arslanBenchmarkDatasetCheckWorthy2020,gangireddyNewsClaimsNewBenchmark2022}, and unverifiable subjective statements are usually not considered as factual claims. In contrast, in argument mining, claim detection aims at identifying the core argument or point of view referring to what is being argued about \citep{habernal-gurevych-2017-argumentation}. Therefore, both objective and subjective information can be identified as claims depending on their role in the discourse \citep{daxenberger-etal-2017-essence,chakrabarty-etal-2019-imho}. The intermixing of such concepts has led to dataset misuse issues in research: for instance, \citet{guptaLESALinguisticEncapsulation2021} annotate a claim detection dataset for fack-checking COVID-19 tweets. However, the dataset is jointly trained and evaluated with claim detection datasets for argument mining \citep[][inter alia]{peldszus-stede-2015-joint,stabParsingArgumentationStructures2017}, which potentially harms the soundness of the results. 

\myparagraph{Discrepancies in task definitions:}
Some prior work defines factual claim detection as identifying check-worthy claims \citep{arslanBenchmarkDatasetCheckWorthy2020,nakov2021overview,nakovOverviewCLEF2022CheckThata,stammbachEnvironmentalClaimDetection2023} while others aim at distinguishing factual claims and non-claims \citep{konstantinovskiyAutomatedFactcheckingDeveloping2020,guptaLESALinguisticEncapsulation2021}. \citet{alam-etal-2021-fighting-covid} and \citet{arslanBenchmarkDatasetCheckWorthy2020} have both check-worthiness and claim vs non-claim labels.
However, \citet{konstantinovskiyAutomatedFactcheckingDeveloping2020} posits that the definition of check-worthiness is subjective, depending on an annotator's knowledge or political stance about a topic. For example, the statement ``human-induced climate change is an immediate and severe threat'' might be deemed self-evident by climate scientists but as checkworthy by others who are skeptical of climate models or prioritize economic growth. Some might argue that claims like this, which are subject to disagreement regarding their importance, are check-worthy due to their controversial nature. However, it requires background knowledge outside the claim itself to determine the controversy. This could involve factors such as who made the claim and why it is controversial, making the task impossible to solve at the sentence level.

Check-worthiness labels also suffer from another serious problem of future prediction. Training a model detecting past check-worthy claims (e.g., about COVID-19) may fail to detect check-worthiness in future claims whose sociopolitical context and controversy are unknown. 

\myparagraph{Blurry boundaries between factual claims and non-claims:}
In related work, personal opinions are usually defined as non-factual claims \citep{arslanBenchmarkDatasetCheckWorthy2020,alam-etal-2021-fighting-covid}. However, many opinions are explicitly based on verifiable facts, lying between the definition of factual claims and non-factual claims. For example: ``Hydroxychloroquine cures COVID.'' is a verifiable factual claim. But ``I believe Hydroxychloroquine cures COVID.'' becomes a personal opinion based on a verifiable fact. \citet{alam-etal-2021-fighting-covid} excludes all opinions from factual claims, which is not a good practice. A false claim can be harmful in political speeches and social media, no matter if it is enclosed by "I believe" or not. \citet{guptaLESALinguisticEncapsulation2021} defines `opinions with societal implications as factual claims'', where societal implications is again an ambiguous definition.

The first row of \cref{tab:examples} showcases the prevalent entanglement of subjective and objective information. To the best of our knowledge, no previous work in factual claim detection discusses the intersection of opinions and facts and how to delineate facts from opinions.

\myparagraph{Context Unavailable:} 
Related work focusing on sentence-level factual claim detection in political speech fails to discuss that sometimes sentences are not self-contained \citep{arslanBenchmarkDatasetCheckWorthy2020,checkthat2023}. However, resolving the co-references is essential for semantic understanding. The last row of \cref{tab:examples} shows such an example.


\subsection{Our Definition of Factual Claims} \label{sec:our_def}
\begin{figure*}[ht!]
	\centering
	\includegraphics[width=1\textwidth]{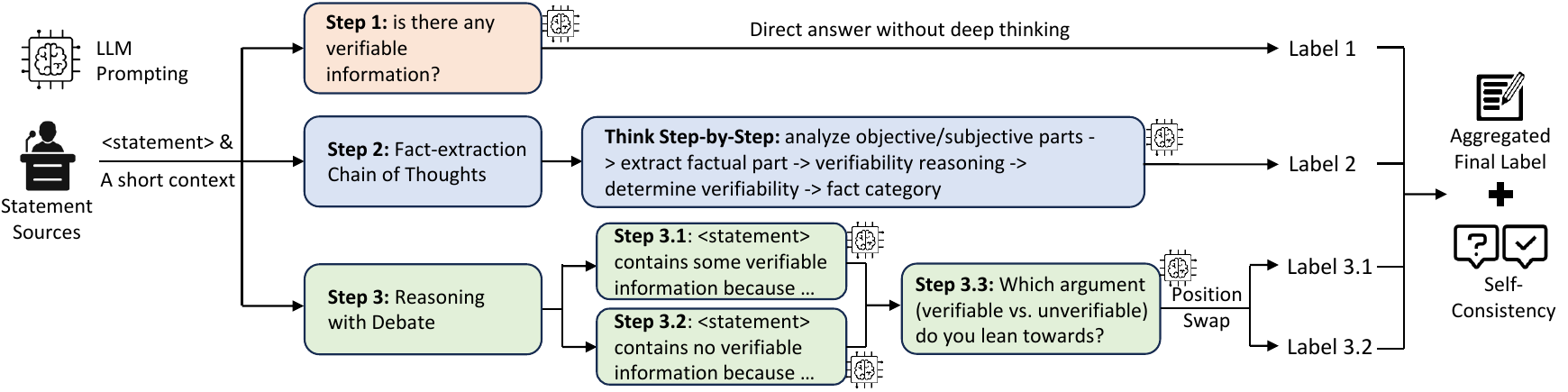}
	\caption{AFaCTA Pipeline. All steps that need LLM prompting are annotated with the brain icon. Besides the target statement, a short context (if available) is also provided to help the model understand the statement.}
	\label{fig:autofactdelineator}
\end{figure*}

To avoid \textbf{claim misconceptions}, we always use ``factual claim'' or ``claim detection for fact-checking'' to specify our focus on fact-checking rather than argument mining. We define facts focusing on verifiability following \citet{arslanBenchmarkDatasetCheckWorthy2020} and \citet{alam-etal-2021-fighting-covid}:
\begin{description}
  \item[Fact:] \textit{A fact is a statement or assertion that can be objectively verified as true or false based on empirical evidence or reality.}
\end{description}

To have \textbf{a clear and objective task definition}, we follow \citet{konstantinovskiyAutomatedFactcheckingDeveloping2020} to focus on verifiability (factual vs. not factual claim) instead of check-worthiness (check-worthy vs. not check-worthy). Whether a sentence contains a verifiable fact or not depends only on its content (and sometimes on a little context surrounding it to clarify key statements), regardless of political or social contexts not captured by the text itself. This differs from many related works that annotate political opinions without verifiable facts as check-worthy and verifiable facts as not check-worthy. Examples of differences in checkworthiness and verifiability are showcased in rows two and three of \cref{tab:examples}. Controversial political opinions and interpretations are usually considered check-worthy due to their potential societal implications. However, they are often open to debate and can hardly be verified against certain evidence. Therefore, we argue that checkworthiness and verifiability are perpendicular dimensions of factual claim detection. In this work, we focus on verifiability for the scalability of data annotation and transferability to easy-to-deploy smaller models.

To address the \textbf{opinion-with-fact problem} that is overlooked by prior work, we define opinions and factual claims as:
\begin{description}
    \item[Opinion:] \textit{An opinion is a judgment based on facts, an attempt to draw a reasonable conclusion from factual evidence. While the underlying facts can be verified, the derived opinion remains subjective and is not universally verifiable.}
    \item [Factual claim:] \textit{A factual claim is a statement that explicitly presents some verifiable facts. Statements with subjective components like opinions can also be factual claims if they explicitly present objectively verifiable facts.}
\end{description}


\myparagraph{How to define verifiability?} 
The verifiability of information is not trivial to define because many assertions can be interpreted either subjectively or objectively. For instance, ``MIT is one of the best universities in the world'' can be either expressing the speaker's subjective feeling about MIT, which is not verifiable, or it can be asserting a verifiable fact, which can be checked with evidence like university rankings and public survey results. For clarity, we define a statement as verifiable if \textbf{it provides enough specific information to guide fact-checkers in verification}. Therefore, the above MIT claim is verifiable. Generally, we observe that a statement is verifiable when it provides specific details for evidence search. For example, ``MIT is a good university'' is less verifiable than ``MIT is one of the best universities according to the QS ranking''.

\section{AFaCTA}

This section introduces AFaCTA for assisting factual claim annotation. AFaCTA consists of three prompting steps and an aggregation step (illustrated in \cref{fig:autofactdelineator}), inspired by \citet{Kahneman2011ThinkingFA} and our claim definitions. The prompts can be found in \cref{app:prompt}. 

\myparagraph{Step 1: Direct Classification.}
We ask LLMs to answer whether a statement contains verifiable information without any chain of thought (CoT, \citealp{wangSelfConsistencyImprovesChain2023}). This step corresponds to a human expert's fast decision-making at first sight of a statement without deep thinking. 
\myparagraph{Step 2: Fact-Extraction CoT.} 
We instruct LLMs to conduct step-by-step reasoning over a statement: firstly, analyze the objective and subjective information covered; secondly, extract the factual part; thirdly, reason why it is verifiable or unverifiable; and finally, determine whether the factual part is verifiable. 
This step aims at identifying verifiable facts entangled with subjective opinions (row 1 of \cref{tab:examples}). The prompt and an illustrative example of this step can be found in \cref{app:step_2}.
\myparagraph{Step 3: Reasoning with Debate.} 
We note that the verifiability of many statements depends on their interpretation. 
Ambiguity between verifiable and unverifiable statements often arises from a lack of specificity, as shown in the examples in \cref{app:vague_samples}.

Imitating a critical thinking process, we first prompt LLMs to argue that the statement contains some (or does not contain any) verifiable information. Then we pass the debating arguments to another LLM call to judge which aspect it leans towards. To address the position bias of LLM-as-a-judge \citep{zhengJudgingLLMasajudgeMTBench2023}, we prompt the final judging step twice, each time with the positions of the verifiable and unverifiable arguments swapped. The prompts and an illustrative example of this step can be found in \cref{app:step_3}.

\myparagraph{Final Step: Results Aggregation.}
We aggregate the results of three steps through majority voting. Labels from steps 1 and 2 each contribute one vote, while two position-swapped labels from step 3 contribute 0.5 votes apiece (3 votes in total). Samples with more than 1.5 votes are classified as positive samples (factual claims), and others as negative samples. See \cref{app:tie_break} for a discussion on tie-breaking. Idealy, if all steps have perfect consistency (0 or 3 votes), the annotation accuracy should be high. 



\section{PoliClaim Dataset}
\begin{table}[t]
\small
\centering
\resizebox{\columnwidth}{!}{
\begin{tabular}{lcccc}
\hline
\textbf{Dataset}             & \textbf{|Sample|} & \textbf{|Claim|}  & \textbf{Supervision} & \textbf{Split}                \\ \hline
PoliClaim$_{test}$  & $816$ & $521$ & $100\%$   &   Test   \\
CheckThat!-2021-dev & $140$ & $114$ & $100\%$ & Test \\
PoliClaim$_{gold}$  & $1953$ & $1154$  & $53\%$  &  Train  \\
PoliClaim$_{silver}$ & $4336$ & $2959$ & $0\%$ & Train \\
PoliClaim$_{bronze}$ & $5320$ & $2661$ & $0\%$ & Train \\ \hline
\end{tabular}
}
\caption{\label{tab:datastat} |Sample| and |Claim| indicate the numbers of samples and positive samples. \textbf{Supervision} indicates the portion of the labels with human supervision. \textbf{Split} indicates if the dataset is used for training or test.}
\end{table}

We obtain a large political speech data from \citet{Picard2022PoliticalMI}, which mainly consists of State of the State (SOTS) speeches (already cleaned and split into sentences). These speeches are governors' major public addresses of the year, thus including meaningful political topics. We randomly sample two speeches from each year, from 1998 to 2021, as training data and four speeches from 2022 as test data.\footnote{We do speech-level random sampling to keep the sentence distribution of full speeches.} This design has two considerations: (1) We aim to replicate the real-world scenario where models are trained on previous claims (e.g., from 1998 to 2021) and used to predict future claims on potentially unseen topics (e.g., in 2022). (2) The test set will be used to evaluate the annotation performance of AFaCTA, and the 2022 speeches are likely unseen by June LLM checkpoints we use to better replicate the future-claim-detection scenario.

The PoliClaim test set (PoliClaim$_{test}$) was annotated by two human experts\footnote{PhD students who are familiar with the domain of political speeches in the U.S. and COVID-related claims and have good knowledge of the literature on claim detection.}, who had no access to AFaCTA's output when annotating. The experts achieved a substantial Cohen's Kappa of 0.69 in independent annotation before the discussion. Then, they had meetings to resolve disagreements and develop gold labels. Disagreements were mainly caused by ambiguous verifiability, see \cref{app:vague_samples} for disagreement resolving. Our annotation guideline, an instantiation of our factual claim definition, can be found in \cref{app:annotation_guideline}. 

To test AFaCTA's annotation performance on different domains, we re-annotate the development set of CheckThat!-2021 \citep{nakov2021overview}, which originally contained check-worthiness labels of COVID-19 tweets, following the same annotation process (Cohen's Kappa 0.58). Due to budget limitations, our explorations and annotations mainly focused on the domain of political speech. We leave the extensive study on the social media domain (and other potential domains for factual claim detection) to future work.

After verifying the performance of AFaCTA using the test sets (see more in \cref{sec:major_result}), we annotated the training set with the tool's assistance, imitating its expected use case of assisting annotation. The perfectly consistent samples were labeled directly with GPT-4 AFaCTA, while the inconsistent samples were left for human annotation. We randomly sampled 8 speeches and manually re-labeled the inconsistent annotations from AFaCTA, leading to PoliClaim$_{gold}$ where all annotations are labeled with perfect consistency or human supervision. The perfectly consistent samples in the rest of the speeches fall into PoliClaim$_{silver}$ while the inconsistent samples fall into PoliClaim$_{bronze}$. The statistics of datasets can be found in \cref{tab:datastat}.

\section{Experiments} \label{sec:experiments}
\begin{table*}[ht]
\small
\centering
\resizebox{\textwidth}{!}{
\begin{tabular}{lcccccccc}
\hline
& \multicolumn{2}{c}{$\mathbf{S}$ ($100^\dagger$/$100^\ddagger$)} & & \multicolumn{2}{c}{$\mathbf{S^{\mathcal{M}}_{con}}$ ($43.38^\dagger$/$48.78^\ddagger$)} & & \multicolumn{2}{c}{\textbf{$\mathbf{S^{\mathcal{M}}_{inc}}$ ($56.62^\dagger$/$51.22^\ddagger$)}} \\
\cline{2-3} \cline{5-6} \cline{8-9}
& \textbf{Agreement} & \textbf{Accuracy} & & \textbf{Agreement} & \textbf{Accuracy} & & \textbf{Agreement} & \textbf{Accuracy} \\
\hline
GPT-3.5 & $0.510$ & $76.47$ & & $0.754$ & $90.40$ & & $0.331$ & $65.80$ \\
GPT-4 & $0.615$ & $86.27$ & & $\mathbf{0.833}$ & $\mathbf{98.49}$ & & $0.418$ & $74.64$ \\
Experts & $\mathbf{0.690}$ & $\mathbf{92.77}$ & & $0.746^\dagger$/$0.743^\ddagger$ & $93.79^\dagger$/$94.85^\ddagger$ & & $\mathbf{0.636}^\dagger$/$\mathbf{0.629}^\ddagger$ & $\mathbf{91.99}^\dagger$/$\mathbf{90.79}^\ddagger$ \\
\hline
\end{tabular}
}
\caption{AFaCTA's performance on PoliClaim$_{test}$. ``$S$'', ``$S^{\mathcal{M}}_{con}$'', and ``$S^{\mathcal{M}}_{inc}$'' report scores on the full test set, perfectly consistent samples, and inconsistent samples correspondingly. The percentages (\%) of ``$S^{\mathcal{M}}_{con}$'' and ``$S^{\mathcal{M}}_{inc}$'' samples are also reported in column titles. The \textbf{Experts} row reports inter-human agreement and average human annotation accuracy against gold labels. \textbf{GPT-3.5 (-4)} rows report AFaCTA's average agreement to both experts, and its accuracy score against gold labels. ``$\dagger$'' and ``$\ddagger$'' denote GPT-3.5 and GPT-4 reported $S^{\mathcal{M}}_{con}$ / $S^{\mathcal{M}}_{inc}$ correspondingly (i.e., $\mathcal{M}=$ GPT-3.5 / -4).}
\label{tab:annotation_comparison}
\end{table*}
Since AFaCTA is an LLM-agnostic prompting framework, we test both GPT-3.5  \citep{ouyang-etal-2021-knowledge-representation} and GPT-4 \citep{openaiGPT4TechnicalReport2023} as the backbone LLM. We also test open-sourced LLMs which does not work well due to high position bias in Step 3 (see \cref{app:open_sourced_llm}). Detailed settings are in \cref{app:hyperparameter} to ensure reproducibility.


\subsection{AFaCTA Annotation Performance} \label{sec:major_result}
It is unlikely for LLMs to produce expert-level annotation on all samples $S$. Therefore, AFaCTA (with LLM $\mathcal{M}$) calibrates its performance with self-consistency, dividing $S$ into two subsets: $S^{\mathcal{M}}_{con}$ with perfect consistency across all steps (0 or 3 votes) and $S^{\mathcal{M}}_{inc}$ with inconsistency among some steps (0.5 to 2.5 votes). We use two criteria to compare AFaCTA with human experts: (1) Accuracy: AFaCTA's accuracy vs. experts' average accuracy, both are computed against gold labels; (2) Agreement (Cohen's Kappa): AFaCTA's average agreement to experts vs. agreement between experts. Both metrics should be compared on $S$, $S^{\mathcal{M}}_{con}$, and $S^{\mathcal{M}}_{inc}$ to evaluate AFaCTA's reliability on entire, perfectly consistent, and inconsistent samples. See \cref{app:formula} for formulas and implementations of all metrics.

The results are presented in \cref{tab:annotation_comparison}. On the full test set $S$, even GPT-4 AFaCTA underperforms the average performance of human experts on both accuracy and agreement. However, if we only consider the subset where AFaCTA has perfect consistency ($S^\mathcal{M}_{con}$), GPT-4 outperforms human experts by a large margin on accuracy (98.49\% > 94.85\%) and achieves better agreement with experts (0.833 > 0.743). 
On the contrary, LLMs achieve worse annotation performance than human experts on inconsistent subsets ($S^\mathcal{M}_{inc}$). Comparable inter-human agreement is achieved on both subsets, but the accuracy and agreement on $S^\mathcal{M}_{con}$ are higher, indicating that $S^\mathcal{M}_{con}$ is slightly less challenging than $S^\mathcal{M}_{inc}$. 

\myparagraph{\textbf{Takeaway}}: With AFaCTA's self-consistency calibration, auto-annotation of perfectly consistent samples can be reliably adopted to reduce manual effort (also see \cref{sec:useful_data}). In the case of PoliClaim$_{test}$, only 51.22\% needs further supervision, while 48.78\% of manual effort is saved with GPT-4-AFaCTA.

\subsection{Error Analysis}
Annotation errors in the fact-checking domain may lead to downstream model inaccuracies. Therefore, we also analyze AFaCTA's errors within the perfectly consistent samples. 
We find that GPT-4 AFaCTA makes false positive errors due to over-sensitivity to granular or implicit facts. It makes false negative errors due to context limitations. GPT-3.5 seems less capable of identifying implicit facts within opinions compared to GPT-4. It sometimes fails to identify facts that are specific enough for verification and asks for more ``specific details''. Roughly $97\%$ of its errors are false negatives caused by misunderstanding verifiability and other hallucinations, indicating that its positive predictions are more reliable.


In \cref{app:error_analyses}, we analyze all errors rather than provide isolated examples to avoid cherry-picking. We hope that this thorough analysis can benefit future research in manual/automatic annotation about factual claims.

\subsection{Predefined Reasoning Paths Matter} \label{sec:predefined}

\begin{figure}[t]
    \centering
    \includegraphics[width=\columnwidth]{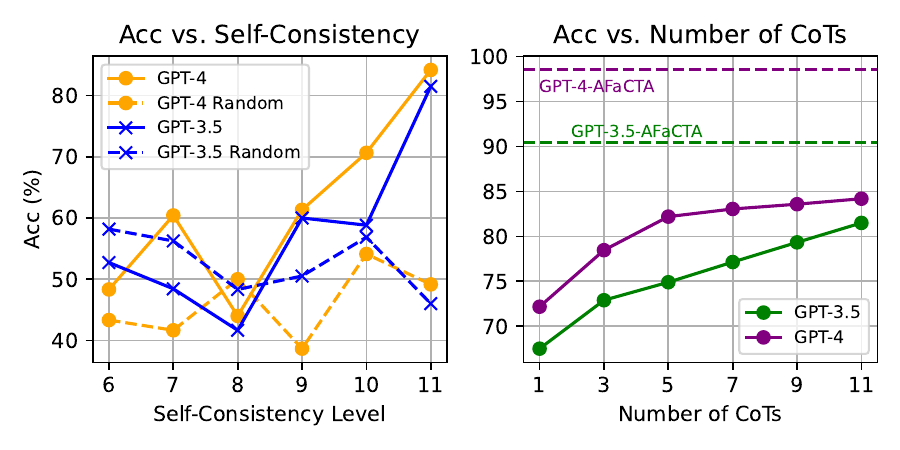}
    \caption{Left figure: accuracy vs. self-consistency levels achieved by $11$ CoT calls. Self-consistency level $x$ means there are $x$ CoTs that agree on the label and $(11-x)$ disagree. Solid and dashed lines denote the performance of LLMs and random guesses on subsets of different self-consistency correspondingly. Right figure: accuracy on the subset where all $x$ CoTs achieve agreement vs. number of sampled CoTs $x$. Note that the subset of perfect consistency is getting narrower and narrower when sampling more CoTs.}
    \label{fig:self-consistency_cot}
\end{figure}
Leveraging self-consistency to improve LLM reasoning is not new. \citet{wangSelfConsistencyImprovesChain2023} show that LLMs can use self-sampled reasoning paths (i.e., CoTs) to improve predictions with self-consistency. In AFaCTA, we use pre-defined reasoning paths instead of LLM-sampled ones. To compare these approaches, we conduct self-consistency CoT with the prompt of Step 1: Direct Classification. Step 1 is chosen since it (1) directly addresses verifiability, which is the core of our factual claim definition; (2) contains no predefined CoT; and (3) is simple but achieves decent performance compared to Steps 2 and 3 (see \cref{app:single_step} where we separately evaluate each step's performance). 

We generate 11 CoTs (more details in \cref{app:cot_prompt}) for both GPT-3.5 and GPT-4 and then compute accuracy scores for different self-consistency levels. The results are illustrated in the left figure of \cref{fig:self-consistency_cot}. We observe that self-consistency level, to some degree, calibrates accuracy: a higher self-consistency level generally indicates higher accuracy, and vice versa. However, self-consistency CoT underperforms AFaCTA on the perfectly consistent subset (84.18\% < 98.49\%) while the former samples 11 CoT reasoning paths, and the latter relies on only 3 predefined reasoning paths. One possible explanation is that the predefined paths encourage critical thinking and reasoning from different angles, making the achieved self-consistency more comprehensive. We also observe that AFaCTA and self-consistency CoT achieve perfect consistency on \textbf{48.78\%} and \textbf{58.09\%} of the data, respectively, indicating that the perfect-consistency in AFaCTA is only slightly harder to achieve than in self-consistency CoT. 

Furthermore, we find that the accuracy on perfectly consistent samples grows with the number of CoT voters (see the right figure of \cref{fig:self-consistency_cot}). This is intuitive as more consistent outputs indicate more confident predictions. However, the marginal benefit of adding more CoTs drops significantly: the accuracy of GPT-4 tends to converge to 85\%. Since the accuracy of GPT-3.5 seems to grow linearly up to 11 CoTs, we further extend it to 19 CoTs and observe convergence to 84.1\% (see \cref{fig:gpt-3.5-extension}), which is still much lower than GPT-3.5 AFaCTA's 90.4\%. 

\myparagraph{\textbf{Takeaway}}: Auto-annotations with more self-consistency (especially the perfectly consistent ones) tend to be more accurate. However, the source of self-consistency needs to be diversified and well-defined to scale up annotation performance efficiently. In this case, we show that predefined reasoning paths with expertise outperform those automatically sampled by LLMs. 

\subsection{Domain Agnostic AFaCTA}
The reasoning logic of AFaCTA is not restricted to the political speech domain. To verify its performance on the social media domain, we conduct the analyses in \cref{sec:major_result} and \cref{sec:predefined} again on the CheckThat!-2021 \citep{nakov2021overview} development set. Experiment results are similar to those on PoliClaim$_{test}$ (see \cref{app:tweet}). Therefore, AFaCTA may assist factual claim annotation in various domains.

\subsection{AFaCTA Delivers Useful Annotations} \label{sec:useful_data}
\begin{figure}[t]
    \centering
    \includegraphics[width=\columnwidth]{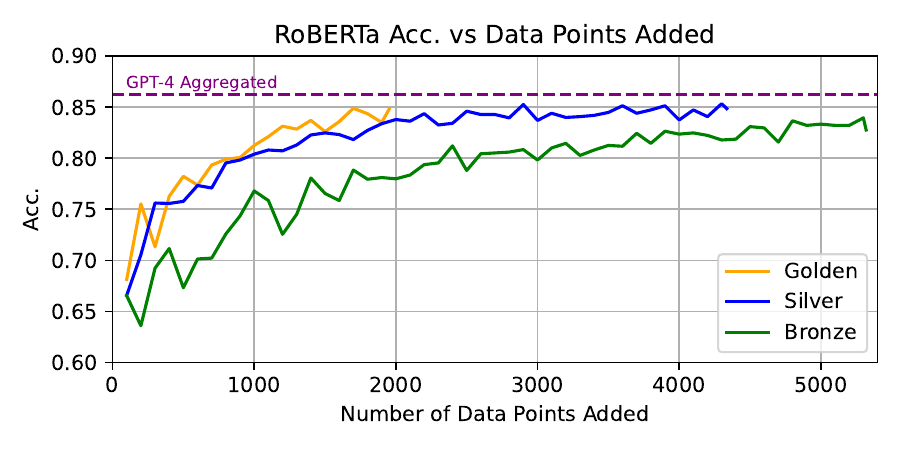}
    \caption{The performance of fine-tuned RoBERTa on PoliClaim$_{test}$ when gradually adding training data of different quality. ``\textcolor{purple}{- -}'' denotes GPT-4's performance aggregating three AFaCTA reasoning steps.}
    \label{fig:pure_roberta}
\end{figure}

\begin{figure*}[t]
    \centering
    \includegraphics[width=\linewidth]{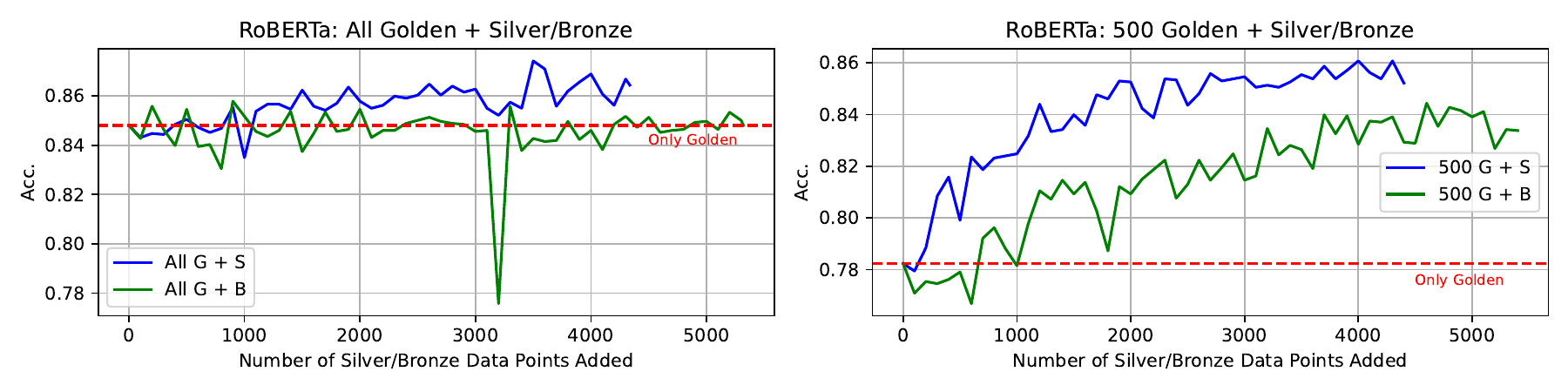}
    \caption{The performance of augmenting a limited number of PoliClaim$_{gold}$ data (left figure: all 1936 samples, right figure: 500 samples) with extra data from PoliClaim$_{silver}$ and PoliClaim$_{bronze}$. Experiments of augmenting 1000 and 1500 PoliClaim$_{gold}$ samples can be found in \cref{app:further_ft}. ``\textcolor{red}{- -}'' denotes the performance without augmentation. G, S, and B denote golden, silver, and bronze PoliClaim correspondingly.}
    \label{fig:add_roberta}
\end{figure*}

To explore whether AFaCTA's annotation can replace or augment manual annotation in training classifiers, we train hundreds of classifiers with different combinations of PoliClaim$_{gold}$ (AFaCTA annotations + Human Supervision), PoliClaim$_{silver}$ (AFaCTA perfectly consistent annotations), and PoliClaim$_{bronze}$ (AFaCTA inconsistent annotations). All results are averaged over random seeds of 42, 43, and 44, and are supported with statistical significance tests (see \cref{app:stat_test}). \footnote{This section presents RoBERTa \citep{roberta} results. \cref{app:further_ft} presents similar DistilBERT \citep{distilbert} results as side findings. Detailed fine-tuning settings are in \cref{app:fine-tuning_settings}.}

\myparagraph{Using only gold, silver, or bronze data:} We first gradually increase the number of training data points (by 100 per step) of the same quality. Results are shown in \cref{fig:pure_roberta}. We observe the same phenomenon as previous work \citep{stammbachEnvironmentalClaimDetection2023} where the marginal accuracy gain drops while adding more data. The PoliClaim$_{gold}$ and PoliClaim$_{silver}$ curves roughly follow the same growing trend, approaching GPT-4's aggregated performance. This indicates that the perfectly consistent annotations (silver) from AFaCTA can strongly substitute for manually annotated data. The PoliClaim$_{gold}$ curve is slightly higher, showing that learning from human-supervised hard samples (inconsistent annotations of AFaCTA) is beneficial. The PoliClaim$_{bronze}$ curve is much lower, showing that the noisy, inconsistent annotations harm the classifier training. 

\myparagraph{Augmenting training with auto-annotated data:} When the manual annotation budget is limited, can we augment the dataset with automatic annotation? In \cref{fig:add_roberta}, we gradually augment the PoliClaim$_{gold}$ data with automatically annotated ones (100 per step). It can be observed that: (1) The performance increases more with PoliClaim$_{silver}$ data augmentation, showing that the data quality is important in data augmentation. (2) Compared to augmenting the full PoliClaim$_{gold}$ dataset, augmentation results in more improvement when there are only 500 PoliClaim$_{gold}$ data. Therefore, high-quality automatic annotation is more helpful when the manual annotation budget is limited. (3) Combining gold and silver data leads to classifiers that outperform aggregated GPT-4 reasoning, demonstrating that extending training data with LLM annotation is a promising approach to achieving better performance. One of the best RoBERTa checkpoints trained on all PoliClaim$_{gold}$ and PoliClaim$_{silver}$ is available on HuggingFace\footnote{https://huggingface.co/JingweiNi/roberta-base-afacta}.




\section{Related Work}
\myparagraph{Claim Detection:}
The term ``claim detection'' has different definitions in various research fields \citep{bolandFactsSurveyConceptualisation2022a}. Even inside the field of fact-checking, its exact definition depends on the domain \citep{alamFightingCOVID19Infodemic2021,stammbachEnvironmentalClaimDetection2023} or task objective \citep{arslanBenchmarkDatasetCheckWorthy2020,konstantinovskiyAutomatedFactcheckingDeveloping2020,gangireddyNewsClaimsNewBenchmark2022} and is somewhat arbitrary. In this work, we propose a definition focusing on one important dimension of factual claims -- verifiability, to minimize the conceptual uncertainty. Another important dimension of factual claims is check-worthiness \citep{arslanBenchmarkDatasetCheckWorthy2020,nakov2021overview,nakovOverviewCLEF2022CheckThata,checkthat2023}, whose definition is more arbitrary \citep{konstantinovskiyAutomatedFactcheckingDeveloping2020}. 



\myparagraph{Automatic Annotation:}
Automatic data annotation using LLM is both promising \citep{pangakisAutomatedAnnotationGenerative2023} and necessary \citep{veselovskyArtificialArtificialArtificial2023}. Early work observes that LLMs' annotation performance highly depends on tasks: LLMs outperform human annotators on some tasks \citep{gilardiChatGPTOutperformsCrowdWorkers2023,zhuCanChatGPTReproduce2023,tornbergChatGPT4OutperformsExperts2023} but fails to achieve human-level performance on others \citep{ziemsCanLargeLanguage2023,reissTestingReliabilityChatGPT2023}. Therefore, we argue that a detailed task-specific study about LLM annotation reliability is essential.

\citet{pangakisAutomatedAnnotationGenerative2023} recommend evaluating LLMs' annotation against a small subset that is not in the LLMs' training corpus and annotated by subject matter experts. We follow these suggestions in this work. Concurrent studies also explore self-consistency \citep{pangakisAutomatedAnnotationGenerative2023} and CoT \citep{heAnnoLLMMakingLarge2023} to improve the performance and reliability of LLM annotation. However, they do not compare predefined reasoning paths with automatically sampled CoTs.

\section{Discussions}
\subsection{Check-Worthiness}
The objective of factual claim detection is to prioritize claims that are both verifiable and check-worthy, maximizing the use of potentially limited fact-checking resources. However, in this project, we focus on verifiability without exploiting the other important aspect: checkworthiness. 
\citet{konstantinovskiyAutomatedFactcheckingDeveloping2020} argues that the definition of check-worthiness is subjective. However, it is possible to define a claim's checkworthiness according to its context. For example, is the claimer an influential person or media? Is the topic controversial? There has already been work that takes some contextual information (e.g., claimer, topic, etc.) into account \citep{gangireddyNewsClaimsNewBenchmark2022}. Future work may explore deterministic and efficient ways to define and annotate checkworthiness leveraging rich contextual information.

\subsection{Only GPT-4 Is Reliable}
We find that only GPT-4-AFaCTA outperforms human experts on perfectly consistent samples. GPT-3.5 achieves promising results but tends to produce false negative errors. Although GPT-4 is much cheaper than human supervision, it is close-sourced and is comparatively more expensive than other LLMs. Future work may study how to use open-sourced models to produce high-quality annotations. Specifically, future work may explore (1) training the model to better understand the annotation guideline; (2) leveraging internal certainties like output logits; and (3) extending the spectrum of self-consistency levels with cheaper inference.

\section{Conclusion}
We propose AFaCTA, which leverages LLMs to assist in the annotation of factual claim detection. It ensures reliability by calibrating annotation quality through consistency. AFaCTA's consistent annotation proves effective for training and data augmentation even without human supervision.


\section*{Limitations}
\myparagraph{AFaCTA Prompt}.
The design of AFaCTA prompts is inspired by the fast and slow thinking patterns \citep{Kahneman2011ThinkingFA} and prior knowledge of factual claim definition. However, we do not explore other techniques (e.g., few-shot prompting, in-context learning, and putting whole annotation guidelines in context etc.) to improve AFaCTA performance further, for two reasons: (1) the current AFaCTA's performance is good enough to show the potential of assisting claim detection annotation with LLMs; and (2) we annotated thousands of sentences with GPT-4-AFaCTA, which is very expensive. Extending the current prompts with more in-context information is not affordable for us.

Besides, AFaCTA step 2 and 3 cost (approximately) 6.5x and 8.5x more tokens than step 1. Although step 2 and 3 bring self-consistency calibration and performance gain through aggregation, the marginal benefit of API cost is far from perfect.

\myparagraph{Social Media and Other Domains}.
In this work, we only conduct extensive experiments and analyses on the political speech domain, only exploring the social media domain with a small dataset (due to the definition discrepancy, we cannot evaluate our methods with prior datasets). We believe a comprehensive study on one domain can provide deeper insights, and the conclusions might be transferable to other domains. Therefore, we do not split our budget across various domains. Future work may consider extending the large-scale analyses to other domains that need fact-checking.

\myparagraph{Limited Expert Annotators}.
We only evaluate AFaCTA's annotation performance against two experts, which may lead to potential bias. We fail to hire more expert annotators mainly because expert annotation is extremely expensive, and it is hard to find more experts with good knowledge about factual claim definitions. As compensation, we release all expert annotations and detailed error analyses where the potential bias can be analyzed. Besides, adding unsupervised LLM-annotated data continuously improves the accuracy on PoliClaim$_{test}$, demonstrating that our human labeling on PoliClaim$_{test}$ has very limited bias.

\section*{Ethics Statement}
In this work, all human annotators are officially hired and have full knowledge of the context and utility of the collected data. We adhered strictly to ethical guidelines, respecting the dignity, rights, safety, and well-being of all participants. 

There are no data privacy issues or bias against certain demographics with regard to the annotated data. Both original SOTS data \citep{Picard2022PoliticalMI} and CheckThat!-2021 \citep{nakov2021overview} datasets are widely used for NLP and other research. Our annotated datasets will also be publicly available for research purpose.

\section*{Acknowledgements} 
\ifarxiv
This paper has received funding from the Swiss
National Science Foundation (SNSF) under the
project `How sustainable is sustainable finance? Impact evaluation and automated greenwashing detection' (Grant Agreement No. 100018\_207800). It is also funded by grant from Hasler Stiftung for the Research Program Responsible AI with the project ``Scientific Claim Verification.''
\fi

\bibliography{anthology,custom}

\appendix

\section{Ambiguities in Verifiability} \label{app:vague_samples}

In political speeches and social media, not all statements are necessarily grounded with enough specific information and are undoubtedly verifiable. Many statements are a mixture of specificity and vagueness, which makes verifiability hard to define. The specificity required for verification may vary based on the topic. But generally, the more specific information a fact contains, the more verifiable it is. For example, a vague statement like "Birmingham is small" tends to be a not verifiable opinion since it lacks specificity (e.g., the standard of ``being small''). In contrast, "Birmingham is small in terms of population compared to London" offers a clearer path for verification by comparing the population sizes of both cities. 
Such ambiguity in verifiability results in different expert annotations. To resolve disagreement and obtain gold labels, we have the experts debate ``whether a statement provides enough specific information to guide fact-checkers in verification'' to achieve agreement.

In the following list, we showcase some examples with vague verifiability. We rely on our experts' critical thinking and common sense to determine their verifiability.

\begin{enumerate}[itemsep=0pt,topsep=1pt,label=E\arabic*.]
    \item \textit{``I promised that our roads would be the envy of the nation.''} Analysis: ``envy of the nation'' seems to be an unverifiable subjective expression. However, this is a part of the speaker's pledge about improving infrastructure and can be verified by comparing the roads with those in other states.
    \item \textit{``Evil acts against innocent people in the places where we once ran errands or recreated have also made us feel less safe.''} Analysis: the speaker claims the existance of evil acts which seems verifiable. However, no specific details are mentioned and different people may interpret or define ``evil act'' differently. Therefore, it is hard to verify.
    \item \textit{``In my budget proposals, we will fully fund our rainy-day accounts.''} Analysis: the "rainy-day account." seems to be an unspecific metaphor which is hard to verify. However, we know from the context that the speaker claims to fund emergency cases (i.e., rainy days). Therefore, it tends to be verifiable.
    \item \textit{``Ensuring society provides a hand up when people need help.''} Analysis: it seems that the speaker is pledging a helpful society. However, nothing specific is mentioned, making this claim hard to verify.
    \item \textit{``Folks, no doubt, the last couple of years have been especially trying for our medical professionals.''} Analysis: at the first glance, the medical professionals' personal feeling seems subjective and not verifiable. However, as COVID is a public event, this can be verified by checking data related to the workload, stress levels, and overal conditions of medical professionals.
    \item \textit{``Authoritarian and illiberal impulses aren’t just rising overseas, they’ve been echoing here at home for some time.''} Analysis: it claims the arising of authoritarian and illiberal impulses. However, no specific events or details are mentioned thus different people may interpret those things differently, making it hard to verify.
    \item \textit{``We are finally going to fix the darn roads.''} Analysis: ``darn roads'' is a subjective expression. However, the speaker's pledge of improving (at least some) roads is verifiable.
    \item \textit{``I’ll call this nonsense what it is, and that is an un-American, outrageous breach of our federal law.''} Analysis: the speaker interprets the COVID vaccination plan as ``an un-American, outrageous breach of federal law'', which seems verifiable by checking laws. However, this is a controversial issue where different people may have different interpretations of the laws. And importantly, no specific legal provisions are mentioned. Therefore, it leans towards unverifiable opinion.
\end{enumerate}

We make all our experts' annotations publicly available. Challenging samples can be found by locating disagreements. Though we tried our best to make the annotation accurate, errors may still occur due to their challenging nature. We encourage future work to improve our definitions to resolve the existing vagueness.

\section{Annotation Guideline} \label{app:annotation_guideline}
The task is to select verifiable statements from political speeches for fact-checking. Given a statement from a political speech and its context, answer two questions following the guidelines. Your annotation will be used to evaluate an LLM-based annotation assistant for factual claim definition.

\subsection{Guidelines}
\myparagraph{Context}: Make sure to consider a small context of the target statement (the previous and next sentence) when annotating. Some statements require context to understand the meaning. For example:
\begin{enumerate}[itemsep=0pt,topsep=1pt,label=E\arabic*.]
    \item ``\textit{... Just consider what we did last year for the middle class in California, sending 12 billion dollars back – the largest state tax rebate in American history. But we didn't stop there. We raised the minimum wage. We increased paid sick leave. Provided more paid family leave. Expanded child care to help working parents ...}'' Without the context, the underlined sentence seems an incomplete sentence. With the context,  we know the speaker is claiming a bunch of verifiable achievements of their administration.
    \item ``\textit{... When I first stood before this chamber three years ago, I declared war on criminals and asked for the Legislature to repeal and replace the catch-and-release policies in SB 91. With the help of many of you, we got it done. Policies do matter. We've seen our overall crime rate decline by 10 percent in 2019 and another 18.5 percent in 2020! ...}'' The underlined part claims that the policies against crimes have been ``done'', which is verifiable. It needs context to understand it.
\end{enumerate}

\myparagraph{Opinion with Facts}: Opinions can also be based on factual information. For example:
\begin{enumerate}[itemsep=0pt,topsep=1pt,label=E\arabic*.]
    \item ``\textit{I am proud to report that on top of the local improvements, the state has administered projects in almost all 67 counties already, and like I said, we've only just begun.}'' The speaker's ``proud of'' is a subjective opinion. However, the content of pride (administered projects) is factual information.
    \item ``\textit{I first want to thank my wife of 34 years, First Lady Rose Dunleavy.}'' The speaker expresses their thankfulness to their wife. However, there is factual information about the first lady’s name and the length of their marriage.
\end{enumerate}

\myparagraph{What is verifiable?} The verifiability of the factual information depends on how specific it is. If there is enough specific information to guide a general fact-checker in checking it, the factual information is verifiable. Otherwise, it is not verifiable. For example:

\begin{enumerate}[itemsep=0pt,topsep=1pt,label=E\arabic*.]
    \item ``\textit{Birmingham is small.}'' is not verifiable because it lacks any specific information for determining veracity. It leans more toward subjective opinion.
    \item ``\textit{Birmingham is small, compared to London}'' is more verifiable than E1. A fact-checker can retrieve the city size, population size ... etc., of London and Birmingham to compare them. However, what to compare to prove Birmingham's ``small'' is not specific enough.
    \item ``\textit{Birmingham is small in population size, compared to London}'' is more verifiable than E1 and E2. A fact-checker now knows it is exactly the population size to be compared.
\end{enumerate}

\myparagraph{When does an opinion explicitly present a fact?} Many opinions are more or less based on some factual information. However, some facts are explicitly presented by the speakers, while others are not. Explicit presentation means the fact is directly entailed by the opinion without extrapolation:

\begin{enumerate}[itemsep=0pt,topsep=1pt,label=E\arabic*.]
    \item ``\textit{The pizza is delicious.}'' This opinion seems to be based on the fact that  ``pizza is a kind of food''. However, this fact is not explicitly presented.
    \item ``\textit{I first want to thank my wife of 34 years, First Lady Rose Dunleavy.}'' The name of the speaker's wife and their year of marriage are explicitly presented.
\end{enumerate}

Along with these guidelines, definitions in \cref{sec:definition} are also presented to the annotators.

\subsection{Annotation Questions}
\myparagraph{Q1. Does the target statement explicitly present any verifiable factual information?} 
\begin{enumerate}[itemsep=0pt,topsep=1pt]
    \renewcommand\labelenumi{$\bullet$}
    \item A - Yes, the statement contains factual information with enough specific details that a fact-checker knows how to verify it. E.g., Birmingham is small in population compared to London.
    \item B - Maybe, the statement seems to contain some factual information. However, there are certain ambiguities (e.g., lack of specificity) making it hard to determine the verifiability. E.g., Birmingham is small compared to London. (lack of details about what standard Birmingham is small)
    \item C - No, the statement contains no verifiable factual information. Even if there is some, it is clearly unverifiable. E.g., Birmingham is small.
\end{enumerate}

\noindent If your answer to Q1 is B - Maybe, then please answer Q2 below:

\myparagraph{Q2. Do you think this statement needs fact-checking of any degree? In other words, does it lean more to checkable facts or subjective opinions?}
\begin{enumerate}[itemsep=0pt,topsep=1pt]
\renewcommand\labelenumi{$\bullet$}
    \item A - Yes, it leans more to facts that need checking.
    \item B - No, it leans more toward subjective opinion and does not need a fact-check.
\end{enumerate}

Samples labeled with A and B/A are positive samples, while those with C and B/B are negative samples.

\section{AFaCTA Prompts} \label{app:prompt}
Following are the prompts of AFaCTA. In all prompts, we always include the previous and next sentence of the target statement if the context is available. ``\{sentence\}'', and ``\{context\}'' are variables to be substituted with the target sentence and its contexts correspondingly. When annotating Twitter data, we simply change ``political speech'' to ``Twitter'' and remove the specifications about contexts (see exact prompts in our code base).

\subsection{System Prompt}
\begin{lstlisting}[frame=single, basicstyle=\ttfamily\scriptsize, numbers=none]
You are an AI assistant who helps fact-checkers to identify fact-like information in statements.
\end{lstlisting}

\subsection{Step 1: Direct Classification}
\begin{lstlisting}[frame=single, basicstyle=\ttfamily\scriptsize, numbers=none]
Given the <context> of the following <sentence> from a political speech, does it contain any objective information? 

<context>: "...{context}..."
<sentence>: "{sentence}" 

Answer with Yes or No only.
\end{lstlisting}

\subsection{Step 2: Fact-Extraction CoT} \label{app:step_2}
In this prompt, we use the categorical definition for facts in \citet{konstantinovskiyAutomatedFactcheckingDeveloping2020}, removing the final category of ``other statements you think are claims'' to reduce uncertainty.

\begin{lstlisting}[frame=single, basicstyle=\ttfamily\scriptsize, numbers=none]
Statements in political speech are usually based on facts to draw reasonable conclusions.

Categories of fact:
C1. Mentioning somebody (including the speaker) did or is doing something specific and objective.
C2. Quoting quantities, statistics, and data.
C3. Claiming a correlation or causation.
C4. Assertion of existing laws or rules of operation.
C5. Pledging a specific future plan or making specific predictions about future.

Please first analyze the objective and subjective information that the following <statement> (from a political speech) covers.
Then extract the fact that the <statement> is based on.
Then carefully reason about if the extracted fact is objectively verifiable. 
Finally answer if the fact falls into the above categories (C1 to C5) or not (C0).

Context for <statement> to help you understand it better: "{context}"
<statement>: "{sentence}"

Format your answer in JSON with the following keys in order: 
{{
    "ANALYSIS": "What are the objective and subjective information that <statement> covers?",
    "FACT_PART": "The extracted fact.",
    "VERIFIABLE_REASON": "Detailed reason about the extracted fact's verifiability. Note that a fact lacks important details or can be interpreted differently is not objectively verifiable. Future plans/pledge (C5) that are specific and clear can be verifiable. Citing others' words is verifiable and falls into C1. ",
    "VERIFIABILITY": "A boolean value indicates the verifiability.",
    "CATEGORY": "C1 to C5, or C0."
}}
\end{lstlisting}

Following is an example showing GPT-4's analyses of a statement, given its context:
\begin{lstlisting}[frame=single, basicstyle=\ttfamily\scriptsize, numbers=none]
Target statement: ``We are so thankful that we haven't suffered any loss of life, and it's always heartening to see and hear stories of Alaskans pitching in to help each other.''

GPT-4's reply: \{
    "ANALYSIS": "The objective information in the statement is that there has been no loss of life due to the storms. The subjective information is the speaker's expression of gratitude and the heartening stories of Alaskans helping each other.",
    "FACT_PART": "There hasn't been any loss of life due to the storms.",
    "VERIFIABLE_REASON": "The fact that there hasn't been any loss of life due to the storms can be verified by checking official records and reports from relevant authorities such as emergency services, hospitals, and local government.",
    "VERIFIABILITY": true,
    "CATEGORY": "C1"
\}
\end{lstlisting}

\subsection{Step 3: Reasoning with Debate} \label{app:step_3}
\noindent Argument for objectivity:
\begin{lstlisting}[frame=single, basicstyle=\ttfamily\scriptsize, numbers=none]
Concisely argue that the following <sentence> from a political speech does contain some objective information.

Context of <sentence> in the speech: "...{context}..."
<sentence>: "{sentence}"
\end{lstlisting}

Following is an example argument from GPT-4 that there is some objective information in the target statement:
\begin{lstlisting}[frame=single, basicstyle=\ttfamily\scriptsize, numbers=none]
Target statement: ``We are so thankful that we haven't suffered any loss of life, and it's always heartening to see and hear stories of Alaskans pitching in to help each other.''

GPT-4's reply: The sentence "The storms have required state disaster declarations covering almost half of our boroughs and the communities they include" does contain objective information. It provides specific details about the impact of the storms, stating that they have led to state disaster declarations in nearly half of the boroughs in the state. This is a factual claim that can be verified by checking the official records of disaster declarations.
\end{lstlisting}

\noindent Argument for subjectivity:
\begin{lstlisting}[frame=single, basicstyle=\ttfamily\scriptsize, numbers=none]
Concisely argue that the following <sentence> from a political speech does not contain any objective information.

Context of <sentence> in the speech: "...{context}..."
<sentence>: "{sentence}"
\end{lstlisting}

Following is an example argument from GPT-4 that there is NO objective information in the target statement:
\begin{lstlisting}[frame=single, basicstyle=\ttfamily\scriptsize, numbers=none]
Target statement: ``We are so thankful that we haven't suffered any loss of life, and it's always heartening to see and hear stories of Alaskans pitching in to help each other.''

GPT-4's reply: The sentence "The storms have required state disaster declarations covering almost half of our boroughs and the communities they include" does not contain objective information because it lacks specific details. It does not provide the exact number or names of the boroughs affected, the specific nature of the disaster declarations, or the precise extent of the damage. Without these details, the statement remains vague and subjective.
\end{lstlisting}

\noindent Final judgment:
\begin{lstlisting}[frame=single, basicstyle=\ttfamily\scriptsize, numbers=none]
Two AI assistants are debating about whether the following <sentence> (from a political speech) contains any objectively verifiable information.

Context of <sentence> in the speech: "...{context}..."
<sentence>: "{sentence}"

Assistant A's View: "{assistant_a}"

Assistant B's View: "{assistant_b}"

Based on the above, does <sentence> contain any objectively verifiable information? Which perspective do you align with more closely? 
Please reply with "Lean towards A", or "Lean towards B" only.
\end{lstlisting}

Following is an example of GPT-4's judge given arguments of different perspectives:
\begin{lstlisting}[frame=single, basicstyle=\ttfamily\scriptsize, numbers=none]
Target statement: ``We are so thankful that we haven't suffered any loss of life, and it's always heartening to see and hear stories of Alaskans pitching in to help each other.'' 

GPT-4's reply: Lean towards A.

Here the objective argument is assigned to assistant A.
\end{lstlisting}

\section{AFaCTA Tie-Breaking} \label{app:tie_break}
All samples with 1.5 votes are labeled as non-factual claims. This tie-break is randomly chosen for two reasons: (1) tie-break does not influence AFaCTA’s performance as it does not affect the perfectly consistent subset (i.e., the subset annotated with best confidence and may not need further human supervision). (2) Samples with 1.5 votes are the most confusing ones reported by AFaCTA, which require human supervision the most. They should not be dogmatically classified in practice. We only do it here to compute AFaCTA's annotation performance on the test sets.

\section{Details of Evaluation Metrics} \label{app:formula}

This section presents the formulas of metrics used in \cref{sec:experiments}. For conciseness, only formulas on perfectly consistent samples $S^{\mathcal{M}}_{con}$ are showcased. Similar formulas are applied for inconsistent samples $S^{\mathcal{M}}_{inc}$ and all samples $S$.

Average accuracy of human expert on perfectly consistent samples $S^{\mathcal{M}}_{con}$ is calculated as:
\begin{equation}
    Acc^{H}_{con} = \!\! \frac{1}{2} \sum_{h \in \{h1, h2\}} \! \! \!  acc\_score(G_{con}, P^{h}_{con})
\end{equation}
where $G_{con}$ and $P^{h}_{con}$ denote the gold labels and human-annotated labels of samples where AFaCTA achieves perfect self-consistency; and $h1$ and $h2$ denotes two human experts.

Accuracy of AFaCTA against gold label on $S^{\mathcal{M}}_{con}$ is calculated as:
\begin{equation}
    \mbox{\it Acc}^{\mathcal{M}}_{con} = \mbox{\it acc\_score}(G_{con}, P^{\mathcal{M}}_{con})
\end{equation}
where $P^{\mathcal{M}}_{con}$ denotes AFaCTA's prediction on perfectly consistent samples. 

Agreement (Cohen's Kappa) between human annotators on $S^{\mathcal{M}}_{con}$ is calculated as:
\begin{equation}
    \mbox{\it Kappa}^{H}_{con} = \mbox{\it cohen\_kappa}(P^{h1}_{con}, P^{h2}_{con}) 
\end{equation}

Average Cohen's Kappa between AFaCTA and two human annotators on $S^{\mathcal{M}}_{con}$ is calculated as:
\begin{equation}
    Acc^{M}_{con} = \frac{1}{2} \!\!\!  \sum_{h \in \{h1, h2\}} \! \! \! \! \! \mbox{\it cohen\_kappa}(P^{h}_{con}, P^{M}_{con}) 
\end{equation}

We use Sci-Kit Learn's accuracy and Cohen's Kappa implementations to calculate all metrics.


\section{AFaCTA with Open-sourced LLMs} \label{app:open_sourced_llm}
\begin{table*}[ht]
\small
\centering
\begin{tabular}{lccccccc}
\hline
& \multicolumn{3}{c}{\textbf{PoliClaim$_{test}$}} & & \multicolumn{3}{c}{\textbf{CheckThat!2021-dev}} \\
\cline{2-4} \cline{6-8}
& \textbf{Agreement} & \textbf{Accuracy} & \textbf{Consistency} & & \textbf{Agreement} & \textbf{Accuracy} & \textbf{Consistency} \\
\hline
zephyr-7b-$\beta$ & 0.205 & 66.18 & 0.49 & & 0.539 & 77.86 & 5.00 \\
llama-2-13b-chat & 0.306 & 56.74 & 0.00 & & 0.260 & 50.71 & 1.43 \\ 
GPT-3.5 & 0.510 & 76.74 & 43.38 & & 0.359 & 69.29 & 44.29 \\
GPT-4 & 0.615 & 86.27 & 48.78 & & 0.437 & 86.43 & 57.85 \\ 
\hline
\end{tabular}
\caption{The performance of AFaCTA with close- and open-source models. We report the average Cohen's Kappa with human experts for agreement, and the accuracy scores are in percentage. We also report the portion of perfectly consistent annotations reported by each model in percentage, which can be found in the consistency column.}
\label{tab:open_source}
\end{table*}

We tried AFaCTA framework on two popular open-sourced LLMs: Llama-2-chat-13b \citep{touvronLlamaOpenFoundation2023} and zephyr-7b-beta \citep{tunstallZephyrDirectDistillation2023}. Results are presented in \cref{tab:open_source}. For both models, we use the official checkpoints on huggingface and conduct greedy decoding when inference. We observe that both models suffer from heavy position bias in AFaCTA step 3: when putting arguments for verifiable and unverifiable to different positions, llama-2-chat-13b and zephyr-7b-beta predict inconsistently in 99\% and 97\% cases correspondingly. Therefore, there are seldom annotations with perfect consistency, and the consistency-based annotation strategy of AFaCTA does not help.

We also observe that zephyr-7b-beta achieves better performance than GPT-3.5 on CheckThat!2021-dev, showing the potential of using open-sourced LLMs as annotators. In future work, we will explore fine-tuning open-sourced LLMs to mitigate the position bias problem and improve annotation quality.

\section{Hyperparameter Settings} \label{app:hyperparameter}
For OpenAI models, we always use gpt-3.5-turbo-0613 and gpt-4-0613. We use a temperature of 0, and top-p of 1 for all experiments except the self-consistency CoT \citep{wangSelfConsistencyImprovesChain2023} experiments where we use a temperature of 0.7. We make all LLM generations publicly available. We always use a random seed of 42 if not specified. For open-sourced LLM inference, we use greedy sampling, a top p of 1, and a maximum generation length of 3072.

\section{Performance of Each AFaCTA Step} \label{app:single_step}
\begin{table*}[ht]
\small
\centering
\begin{tabular}{lcccccccc}
\hline
& \multicolumn{2}{c}{\textbf{Step 1}} & & \multicolumn{2}{c}{\textbf{Step 2}} & & \multicolumn{2}{c}{\textbf{Step 3}} \\
\cline{2-3} \cline{5-6} \cline{8-9}
& \textbf{Agreement} & \textbf{Accuracy} & & \textbf{Agreement} & \textbf{Accuracy} & & \textbf{Agreement} & \textbf{Accuracy} \\
\hline
GPT-3.5 & 0.458 & 73.16 & & 0.452 & 78.06 & & 0.546 & 66.42 \\
GPT-4 & 0.633 & 85.54 & & 0.437 & 79.90 & & 0.630 & 73.28 \\
\hline
\end{tabular}
\caption{The performance of each AFaCTA steps. Similar to \cref{tab:annotation_comparison}, we report the average Cohen's Kappa with human experts for agreement, and the accuracy scores are in percentage.}
\label{tab:each_step}
\end{table*}

We compute the annotation performance of each AFaCTA reasoning step. For Step 3, we average the scores of labels 3.1 and 3.2 (see \cref{fig:autofactdelineator}). The results are presented in \cref{tab:each_step}. It can be observed that Step 1, though simple, achieves promising performance. It outperforms other steps by a wide margin with GPT-4.

\section{Self-Consistency CoT} \label{app:cot_prompt}
We use the following prompt to generate Self-consistency CoT. It keeps most of the prompt template of AFaCTA Step 1 to make them comparable. We use a temperature of 0.7 to sample different CoTs.
\begin{lstlisting}[frame=single, basicstyle=\ttfamily\scriptsize, numbers=none]
Given the <context> of the following <sentence> from a political speech, does it contain any objective information? 

<context>: "...{context}..."
<sentence>: "{sentence}" 

Format your reply as follows:

[Chain of thought]: your step-by-step reasoning about the question
[Answer]: a single word yes or no
\end{lstlisting}

\begin{figure}[t]
    \centering
    \includegraphics[width=\columnwidth]{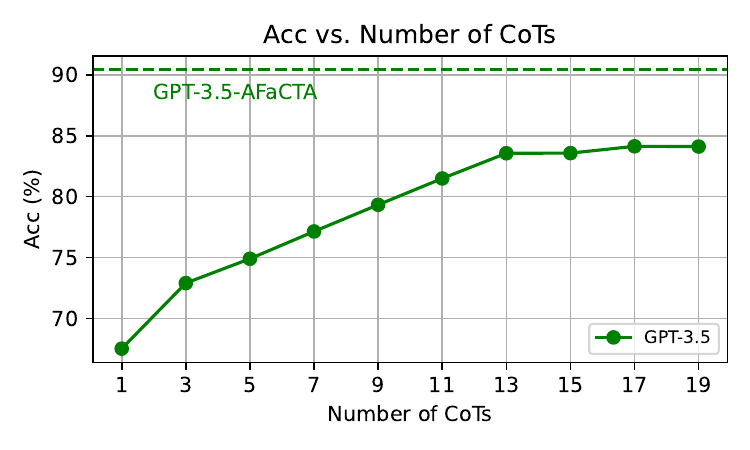}
    \caption{We notice that in \cref{fig:self-consistency_cot}, GPT-3.5's accuracy on the perfectly consistent set does not seem to converge with 11 voters. So we extend the number of CoTs to 19, observing that the accuracy converges to 84.1\%.}
    \label{fig:gpt-3.5-extension}
\end{figure}

\section{Experiments on Social Media Domain} \label{app:tweet}
\begin{table*}[ht]
\small
\centering
\resizebox{\textwidth}{!}{
\begin{tabular}{lcccccccc}
\hline
& \multicolumn{2}{c}{$\mathbf{S}$ ($100^\dagger$/$100^\ddagger$)} & & \multicolumn{2}{c}{$\mathbf{S^{\mathcal{M}}_{con}}$ ($44.29^\dagger$/$57.85^\ddagger$)} & & \multicolumn{2}{c}{\textbf{$\mathbf{S^{\mathcal{M}}_{inc}}$ ($55.71^\dagger$/$42.15^\ddagger$)}} \\
\cline{2-3} \cline{5-6} \cline{8-9}
& \textbf{Agreement} & \textbf{Accuracy} & & \textbf{Agreement} & \textbf{Accuracy} & & \textbf{Agreement} & \textbf{Accuracy} \\
\hline
GPT-3.5 & $0.359$ & $69.29$ & & $\mathbf{0.584}$ & $79.03$ & & $0.205$ & $61.54$ \\
GPT-4 & $0.437$ & $86.43$ & & $0.566$ & $\mathbf{96.30}$ & & $0.280$ & $72.89$ \\
Experts & $\mathbf{0.579}$ & $\mathbf{92.86}$ & & $0.514^\dagger$/$0.540^\ddagger$ & $91.13^\dagger$/$95.68^\ddagger$ & & $\mathbf{0.638}^\dagger$/$\mathbf{0.536}^\ddagger$ & $\mathbf{94.23}^\dagger$/$\mathbf{88.98}^\ddagger$ \\
\hline
\end{tabular}
}
\caption{AFaCTA's performance on our re-annotated CheckThat!-2021-dev. Similar rows, columns, and scores are reported as \cref{tab:annotation_comparison}.}
\label{tab:tweet}
\end{table*}

\begin{figure}[t]
    \centering
    \includegraphics[width=\columnwidth]{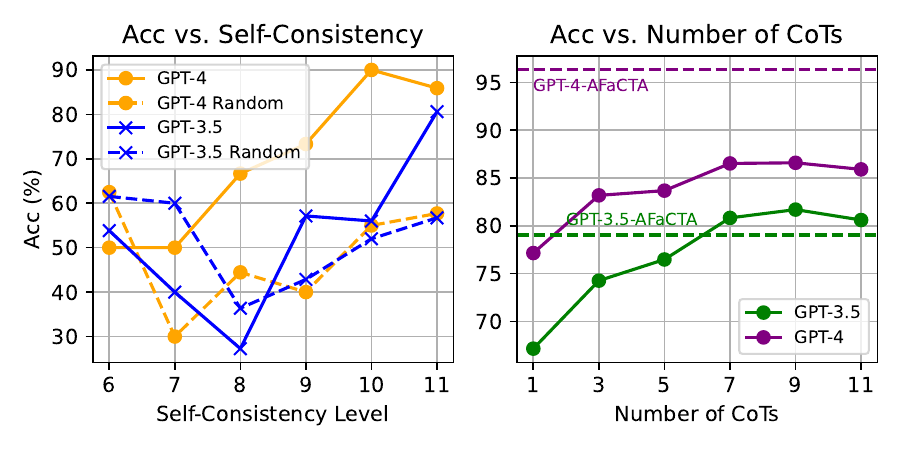}
    \caption{Self-consistency CoT experiments on CheckThat!-2021-dev. Same metrics are reported as \cref{fig:self-consistency_cot}.}
    \label{fig:self-consistency_tweet}
\end{figure}

We compare AFaCTA's annotation performance with human experts on the re-annotated CheckThat!-2021 development set. We have chosen this small set of social media data due to the limitation of the annotation budget.

Similar observations as PoliClaim$_{test}$ can be drawn. GPT-4 AFaCTA outperforms experts on perfectly consistent samples and underperforms on inconsistent samples. GPT-3.5 also achieves a moderate agreement with human experts on perfectly consistent samples. Error analysis shows that GPT-3.5's error concentrates on false negatives, similar to its behavior in the political speech domain (see \cref{tab:t_g35_error}). 

We also conduct the self-consistency CoT experiments on CheckThat!-2021-dev to verify the importance of a diversified source of self-consistency. The results are shown in \cref{fig:self-consistency_tweet}. It can be observed that the level of self-consistency calibrates accuracy, and the 3 predefined reasoning paths outperform automatically generated ones. One discrepancy is that self-consistency CoT slightly outperforms GPT-3.5 AFaCTA when sampling more than 7 reasoning paths. We attribute this to GPT-3.5's heavier hallucinations on Twitter domain (see \cref{tab:t_g35_error} where it fails to identify apparent factual information). Therefore, complicated reasoning paths like AFaCTA Step 3 might be challenging in many cases.

Importantly, due to the annotation budget, our experimental dataset on the social media domain is limited. We leave the extensive analysis of this domain to future work.

\section{Fine-tuning Settings} \label{app:fine-tuning_settings}
For all RoBERTa and DistilBERT fine-tuning experiments, we keep all settings the same except for the training data. All models are fine-tuned for 5 epochs with a batch size of 64. We do not conduct checkpoint selection. For other hyperparameters, we keep the default setting of huggingface TrainingArgument: a learning rate of 5e-5, a max\_grad\_norm of 1, no warm-up and weight decay, etc. We use the huggingface checkpoints of ``roberta-base'' and ``distilbert-base-uncased''. All experiments are conducted on a node with 4 32G V100 GPUs. It takes roughly 0.1 GPU hour to train a classifier. In this work, we always use Sci-kit Learn for score computing.

\section{Statistical Significance Test} \label{app:stat_test}
We conduct a statistical significance test to show that different training set combinations of PoliClaim$_{gold}$, PoliClaim$_{silver}$, and PoliClaim$_{bronze}$ lead to statistically significant differences in fine-tuning claim detectors. We first conduct a Student-t test for each training combination based on the results of three random seeds and then aggregate p-values using Fisher's method. For example, to compare ``only PoliClaim$_{gold}$'' vs. only ``PoliClaim$_{silver}$'', we use the following formula:

\begin{align}
    p_{x00} & = \text{Student-t}(\{Acc^{r}_{x00g}\}, \{Acc^{r}_{x00s}\}) \\
    p_{agg} & = \text{Fisher}(p_{100}, p_{200}, ..., p_{2000})
\end{align}
where $r$ denotes random seeds 42, 43, and 44; $p_{x00}$ denotes the p-value of the x00 step; and $p_{agg}$ denotes the aggregated p-value. The aggregated p-values of all comparisons are shown in \cref{tab:stat_test}. It can be seen that all observations in \cref{sec:useful_data} and \cref{app:further_ft} are statistically significant. Scipy's implementations for Student-t test and Fisher's Method are used.

\begin{table}[t]
\small
\centering
\begin{tabular}{c@{\hspace{3pt}}c@{\hspace{3pt}}ccc}
\toprule
    \multicolumn{3}{c}{\textbf{Comparison}}& \textbf{RoBERTa}  & \textbf{DistilBERT}               \\ \hline
Only S & < & Only G  & 5.54e-3$^{*}$ & 8.89e-5$^{**}$ \\
Only B & < & Only S  & 2.39e-36$^{**}$ & 5.79e-51$^{**}$ \\
Only B & < & Only G  & 1.88e-20$^{**}$ & 6.03e-29$^{**}$ \\
500 G + B & < & 500 G + S  & 1.50e-28$^{**}$ & 1.82e-30$^{**}$ \\
1000 G + B & < & 1000 G + S  & 8.13e-13$^{**}$ & 3.30e-8$^{**}$ \\
1500 G + B & < & 1500 G + S  & 2.19e-16$^{**}$ & 1.69e-15$^{**}$ \\
All G + B & < & All G + S  & 3.68e-9$^{**}$ & 1.36e-13$^{**}$ \\
\bottomrule
\end{tabular}
\caption{\label{tab:stat_test} Statistical significance of performance difference with different train sets. G, S, and B denotes PoliClaim$_{gold}$, PoliClaim$_{silver}$, and PoliClaim$_{bronze}$ correspondingly. By $^{*}$ and $^{**}$, we denote a p-value smaller than $0.01$ and $0.001$, respectively.}
\end{table}

We do not conduct statistical tests on experiments of \cref{sec:major_result} as obtaining independent samples of human / GPT-4 annotation can be very costly, and OpenAI API does not support random seeds at the moment of experimenting.

\section{Further Fine-tuning Experiments} \label{app:further_ft}
This section provides more supplementary results of the experiments in \cref{sec:useful_data}.
\subsection{Only Golen, Silver, or Bronze}
\begin{figure}[t]
    \centering
    \includegraphics[width=\columnwidth]{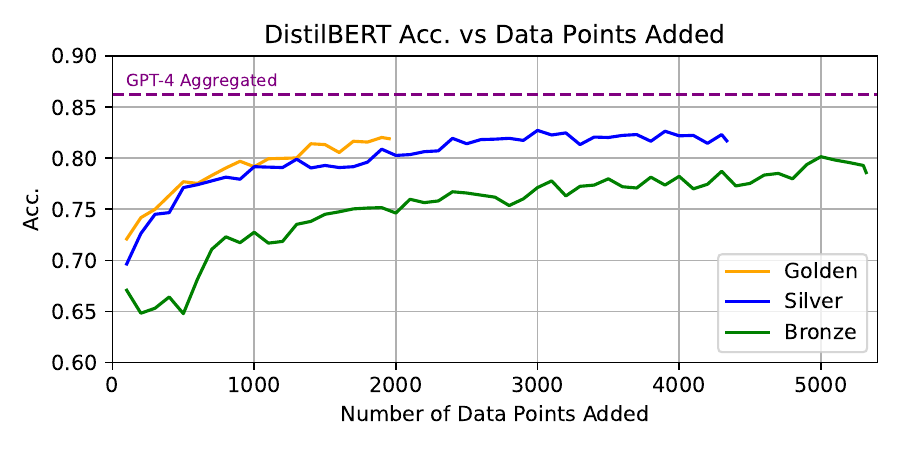}
    \caption{The performance of fine-tuned DistilBERT on PoliClaim$_{test}$ when gradually adding training data of different quality. Same scores are reported as \cref{fig:pure_roberta}.}
    \label{fig:pure_distilbert}
\end{figure}
We gradually increase the size of golden, silver, and bronze training data to fine-tune DistilBERT. The results are shown in \cref{fig:pure_distilbert}. The same observations can be drawn from \cref{fig:pure_roberta}: perfectly consistent (silver) data achieve a similar growing trend as manually supervised (golden) data, while accuracy grows slower when adding (bronze) inconsistent data. 

\subsection{Augmenting Gold Data with Silver/Bronze Data}
\begin{figure*}[t]
    \centering
    \includegraphics[width=\linewidth]{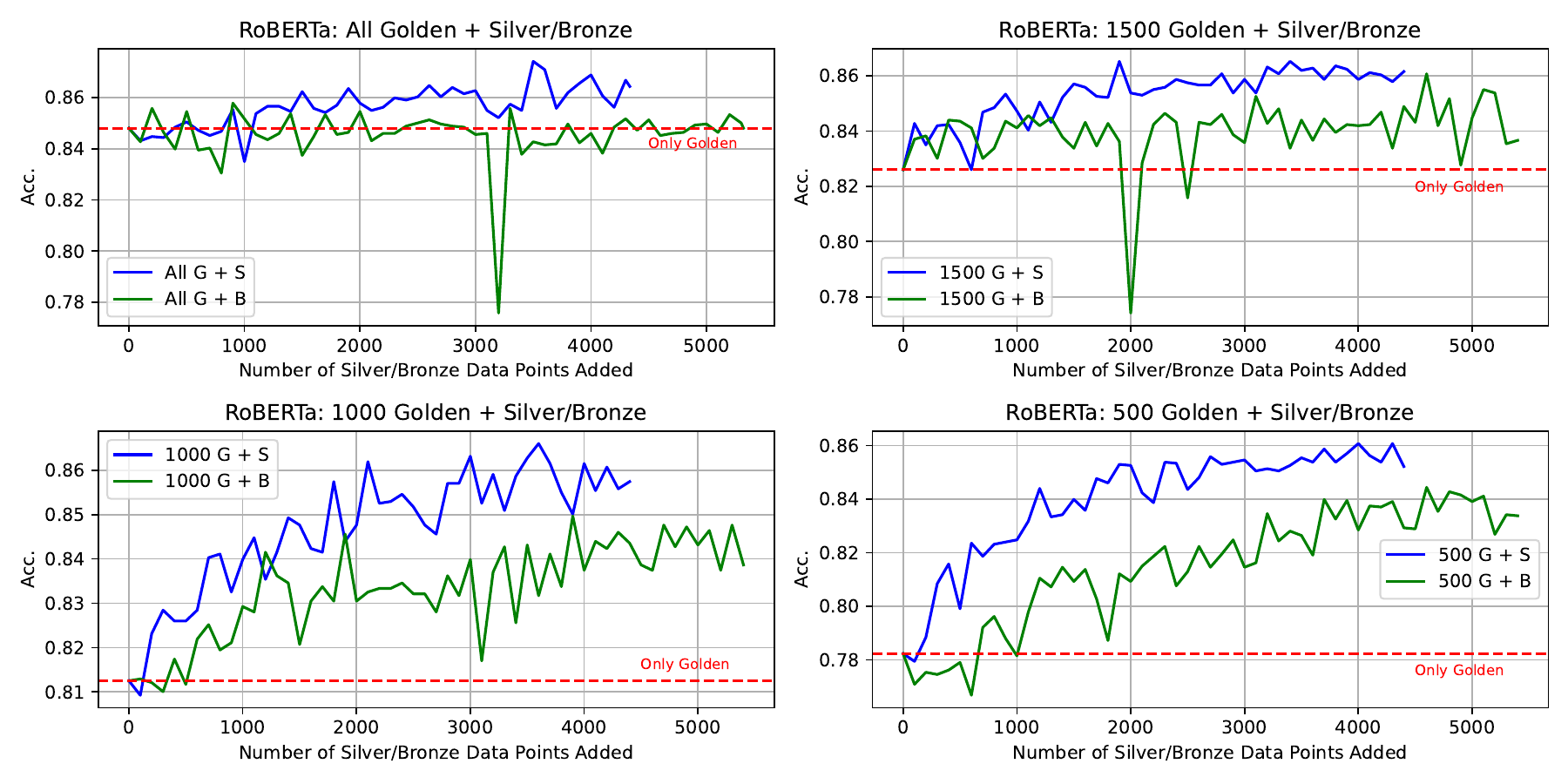}
    \caption{The RoBERTa performance of augmenting a limited number of PoliClaim$_{gold}$ data. An augmented version of \cref{fig:add_roberta} with 1000 and 1500 Gold data experiments added.}
    \label{fig:add_all_roberta}
\end{figure*}

\begin{figure*}[t]
    \centering
    \includegraphics[width=\linewidth]{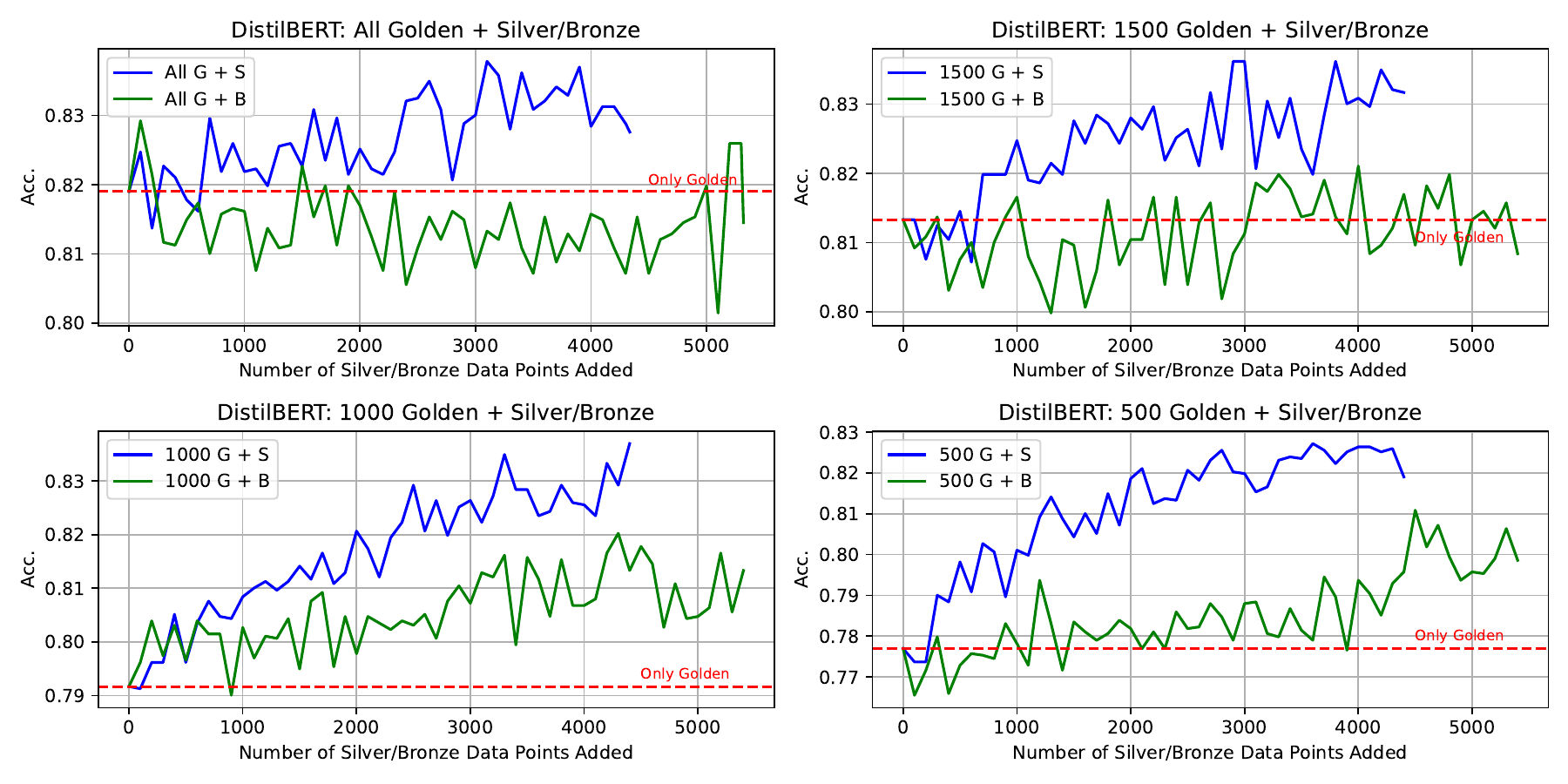}
    \caption{The DistilBERT performance of augmenting a limited number of PoliClaim$_{gold}$ data. The same scores are reported as \cref{fig:add_all_roberta}.}
    \label{fig:add_distilbert}
\end{figure*}

\begin{figure*}[t]
    \centering
    \includegraphics[width=\linewidth]{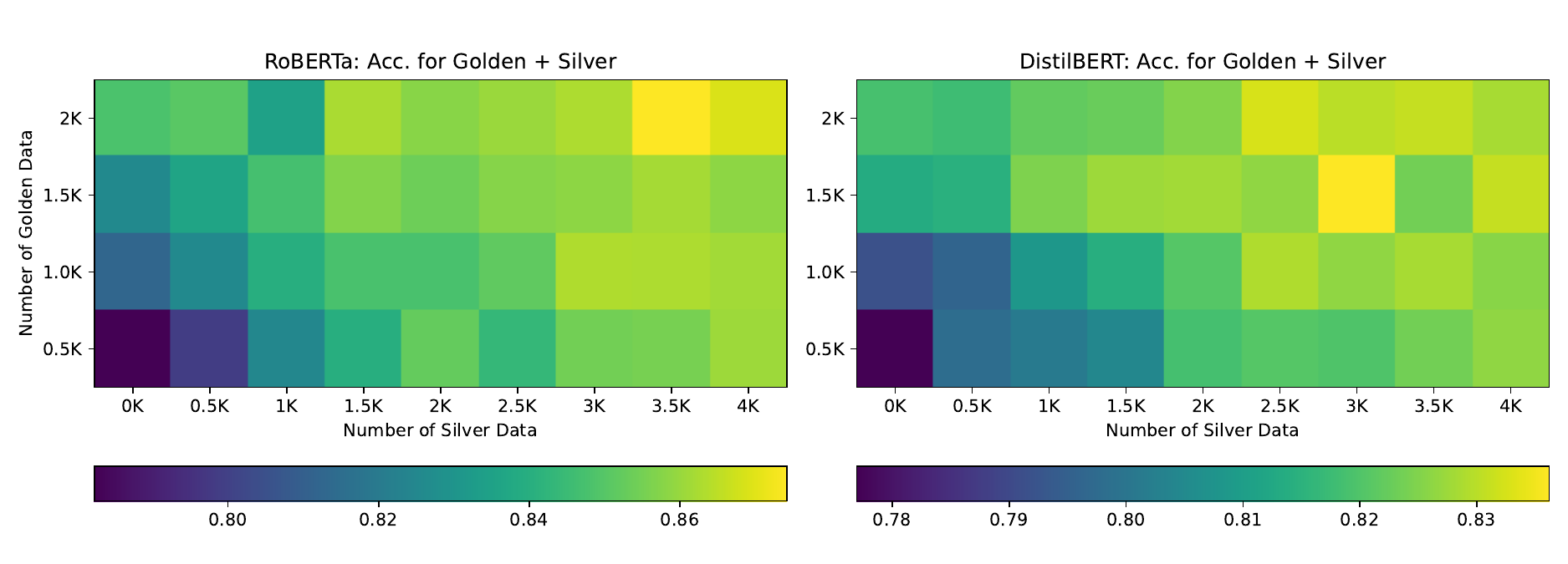}
    \caption{The performance of combining different amount of PoliClaim$_{gold}$ and PoliClaim$_{test}$.}
    \label{fig:colormap}
\end{figure*}

We conduct the data augmentation experiments in \cref{sec:useful_data} on both RoBERTa (\cref{fig:add_all_roberta}) and DistilBERT (\cref{fig:add_distilbert}) with a different number of PoliClaim$_{gold}$ data (500, 1000, 1500, and 1936). Similar conclusions as \cref{sec:useful_data} can be drawn: perfectly consistent (silver) data are better at augmentation than inconsistent (bronze) data. \cref{fig:colormap} also shows a clear trend. When the manual annotation budget is more restricted, more augmentation data are needed to achieve a comparable performance.

In all experiments, the marginal benefit of adding data decreases quicker on DistilBERT than on RoBERTa, as expected. However, we suspect adding more high-quality annotated and diversified data might boost weaker models to outperform stronger models, though the marginal accuracy gain is low. We leave this exploration to future work.

\section{Error Analyses}
\label{app:error_analyses}
\begin{table*}[t]
\small
\centering
\resizebox{\textwidth}{!}{
\begin{tabular}{>{\raggedright\arraybackslash}m{0.13\textwidth}m{0.87\textwidth}}
\hline
\textbf{Error Type} & \multicolumn{1}{c}{\textbf{GPT-4 Errors and Explanations}} \\ \hline

False Positive: over-sensitive to granular, unspecific, or not-explicitly-presented facts & \textit{\textbf{Error 1:}} \hl{I just want to thank you, thank you members of the Legislature for all you did these past two years to keep us safe.} \textit{Error reason: recognizing 'members of the legislature did some thing' as fact, which is too vague.}

\textit{\textbf{Error 2:}} \hl{It's true from the Flatirons to Fishers Peak to Pikes Peak to Longs Peak and beyond.} \textit{Error reason: location names are recognized as facts.}

\textit{\textbf{Error 3:}} \hl{Sheriff Pelle, firefighters, and emergency responders, please stand so we can thank you for the lifesaving work that you do every day.} \textit{Error reason: people's appearance at the event is recognized as a fact, which is not explicitly presented.}

\textit{\textbf{Error 4:}} \hl{I'm glad to be back at the capitol addressing the Legislature in person, and I thank you for the invitation to speak to you tonight.} \textit{Error reason: identify the speaker's back and addressing legislature as facts.}
 \\ \hline
 
False Negative: not enough context & \textit{\textbf{Error 5:}} \hl{It's the result of great investment decisions, policies, vision, and direction.
} \textit{Error reason: ``it'' here refers to the return of pension fund, which is in a far context. But AFaCTA only considers a one-sentence context.}

\textit{\textbf{Error 6:}} \hl{Alaskans won't accept that we can't get anything done because it's an election year.} \textit{Error reason: it claims a fact that this year is an election year, but the model comprehends this as a hypothetical condition, due to its lack of context that 2022 is the election year for Alaska.}
\\ \hline

\end{tabular}
}
\caption{\label{tab:g4_error} All errors made by GPT-4 AFaCTA on PoliClaim$_{test}$. Statements are highlighted in \hl{yellow}. The reasons for making errors are written in \textit{italics}.}
\end{table*}

\begin{table*}[t]
\small
\centering
\resizebox{\textwidth}{!}{
\begin{tabular}{>{\raggedright\arraybackslash}m{0.13\textwidth}m{0.87\textwidth}}
\hline
\textbf{Error Type} & \multicolumn{1}{c}{\textbf{GPT-3.5 Errors and Explanations}} \\ \hline

False Positive: over-sensitive to unspecific facts & \textit{\textbf{Error 1:}} \hl{The fresh mountain air that so many people associate with Colorado isn't a given.} \textit{Error reason: identifying ``the fresh air is not a given'' as a fact, which is unspecific and leans towards unverifiable opinion.}
 \\ \hline
 
False Negative: not enough specific detail or context and thus not verifiable & \textit{\textbf{Error 2:}} \hl{When our federal government overreached, we found a way to fight back.} \textit{Error reason: the model argues it lacks details of ``overreach''.}

\textit{\textbf{Error 3:}} \hl{While our work is far from over, we have made significant progress thanks to the Rebuild Alabama Act.} \textit{Error reason: the model argues it lacks important details of ``significant progress''.}

\textit{\textbf{Error 4:}} \hl{Folks, no doubt, the last couple of years have been especially trying for our medical professionals.} \textit{Error reason: the model argues the ``especially trying'' lacks detail.}

\textit{\textbf{Error 5:}} \hl{I am proud that my Administration, with the support of the Legislature, is doing more to make significant improvements in mental health care than any since Governor Lurleen Wallace in the 1960s.} \textit{Error reason: the model argues the ``significant improvements'' lack detail.}

\textit{\textbf{Error 6:}} \hl{At times, her schoolwork and distance from her home state made her wonder if she should give up her Miss Alaska title.} \textit{Error reason: the model argues the ``significant improvements'' lack detail.}

\textit{\textbf{Error 7:}} \hl{It's the result of great investment decisions, policies, vision, and direction.} \textit{Error reason: not enough context about ``it''.}

\textit{\textbf{Error 8:}} \hl{Together with these partners, we'll build a stronger, more durable health care system in Alaska that can respond to most any situation.} \textit{Error reason: the model argues it lacks details about the plan.}

\textit{\textbf{Error 9:}} \hl{At the same time, our ability to increase production is under attack from Washington, DC, and federal courts that side with extremist environmental groups.} \textit{Error reason: the model argues it lacks details about the ``attack''.}

\textit{\textbf{Error 10:}} \hl{No state has been targeted more by the current administration than our Great State of Alaska.} \textit{Error reason: the model argues it lacks details about specific actions.}

\textit{\textbf{Error 11:}} \hl{No state has been targeted more by the current administration than our Great State of Alaska.} \textit{Error reason: the model argues it lacks details about specific actions.}

\textit{\textbf{Error 12:}} \hl{At every turn and since day one of the Biden Administration, this hostility has been perfectly clear.} \textit{Error reason: the model argues it lacks details about the ``hostility''.}

\textit{\textbf{Error 13:}} \hl{Because no president should have to beg for more oil from the Middle East or Russia's Arctic when we can produce it right here better and safer than anywhere else on the planet! This is common sense!
} \textit{Error reason: the model argues it lacks details or evidence about ``the US produces better oil''.}

\textit{\textbf{Error 13:}} \hl{Many of them have been with us for so long that they've almost been normalized in Alaska, as almost unsolvable.} \textit{Error reason: lacking context and details about ``long issues''.}

\textit{\textbf{Error 14:}} \hl{I will always stand between Alaskans and a federal government that violates our God-given rights and exceeds its constitutional authority.
} \textit{Error reason: the model argues it lacks details about specific actions of the federal government and the speaker's future action.}

\textit{\textbf{Error 15:}} \hl{I envision an Alaska where our cost of energy is no longer the second-highest in the nation, but one of the lowest. That's my vision. I hope it is yours as well.} \textit{Error reason: the model argues it lacks details about the definition of ``second-highest'' and ``lowest''.}

\textit{\textbf{Error 16:}} \hl{I've seen it in the men and women on the frontlines of this pandemic who have helped us achieve one of the shortest shutdowns and one of the lowest death rates in the country.} \textit{Error reason: the model argues it lacks details about ``shortest'' and ``lowest''.}

\textit{\textbf{Error 17:}} \hl{And because we want to lead by example, we are saving Coloradans money by making your State Government more efficient and effective.} \textit{Error reason: the model argues it lacks details about ``efficient'' and ``effective''.}

\textit{\textbf{Error 18:}} \hl{Just as an earthquake is followed by aftershocks, we know that the overarching crisis of the pandemic has led to many other crises, perhaps lesser seen, but no less important to address.} \textit{Error reason: the model argues it lacks details about ``the crises''.}

\textit{\textbf{Error 19:}} \hl{We owe it to the people of Colorado to improve safety and make Colorado truly one of the ten safest states in the nation over the next five years.} \textit{Error reason: the model argues it lacks details about the speaker's plan.}

\textit{\textbf{Error 20:}} \hl{No other place offers opportunity to so many from such diverse backgrounds.
} \textit{Error reason: the model argues it lacks specific details.}

\textit{\textbf{Error 21:}} \hl{It's that, as our businesses grow, we don't leave our workers behind.
} \textit{Error reason: the model argues it lacks specific details about business growth.}

\textit{\textbf{Error 22:}} \hl{By creating choices - real choices - for parents, and unprecedented support for their kids.} \textit{Error reason: the model argues it lacks specific details about the choices and supports.}
\\ \hline
\end{tabular}
}
\caption{\label{tab:g35_error} The only false positive error and the major type of false negative errors made by GPT-3.5 AFaCTA on PoliClaim$_{test}$.}
\end{table*}

\begin{table*}[t]
\small
\centering
\resizebox{\textwidth}{!}{
\begin{tabular}{>{\raggedright\arraybackslash}m{0.13\textwidth}m{0.87\textwidth}}
\hline
\textbf{Error Type} & \multicolumn{1}{c}{\textbf{GPT-3.5 Errors and Explanations}} \\ \hline
False Negative: understand facts as opinions or fail to identify facts entangled with opinions & \textit{\textbf{Error 23:}} \hl{They plan their lives around hunting season, or fishing season; construction season, or tourism season.But not election season.} \textit{Error reason: the model misunderstands it as the speaker's opinion. But people's lifestyles and priorities can be verified with related surveys or studies.} 

\textit{\textbf{Error 24:}} \hl{A future where a dynamic, multi-modal transportation system meets the needs of our growing population.} \textit{Error reason: the model fails to identify ``our growing population'' as a fact.} 

\textit{\textbf{Error 25:}} \hl{When I was elected Governor, I knew that I would be remembered not for who I was, where I came from, or even what I said at events like this, but for what I did to make a meaningful, measurable, positive impact on the lives of Coloradans.} \textit{Error reason: the model fails to identify that ``the speaker is elected as the governor'' is a fact.}

\textit{\textbf{Error 26:}} \hl{But over time, we've learned we can't solve big problems like climate change situationally, with short-term thinking.} \textit{Error reason: the model fails to identify the causality claim about short-term thinking and big problems.}

\textit{\textbf{Error 27:}} \hl{But at a time, when we've been heating and burning up, one thing we cannot do is repeat the mistakes of the past by embracing polluters.} \textit{Error reason: the model fails to recognize the fact of embracing polluters in the past.}
\\ \hline
False Negative: hallucinate about personal experience and citation & \textit{\textbf{Error 28:}} \hl{At times, her schoolwork and distance from her home state made her wonder if she should give up her Miss Alaska title.} \textit{Error reason: the model argues the personal experience is subjective.}

\textit{\textbf{Error 29:}} \hl{``A lot of people,'' she said, ``don't recognize that their low points are what are going to propel them to their future.} \textit{Error reason: subjective personal experience.} 

\textit{\textbf{Error 30:}} \hl{I agree with former Governor Jay Hammond that the government should never take more from the Permanent Fund than is distributed to the people of Alaska.} \textit{Error reason: fail to detect the citation.}

\textit{\textbf{Error 31:}} \hl{She is in a healthy marriage and is reconnecting with her children.} \textit{Error reason: consider personal experience as unverifiable.}

\textit{\textbf{Error 32:}} \hl{``Dad,'' Catherine said, ``Alaska has so much to offer.''} \textit{Error reason: fail to detect the citation.}

\textit{\textbf{Error 33:}} \hl{Still, she found the strength to take down the shooter, ending his violent killing spree and saving many precious lives.} \textit{Error reason: consider personal experience as unverifiable.}

\\ \hline

False Negative: hallucinate about rhetoric & \textit{\textbf{Error 34:}} \hl{They're wondering how we've come to a place where the PFD is nothing more than what's left over after government takes the lion's share.
} \textit{Error reason: fail to understand the metaphor.}
\\ \hline
\end{tabular}
}
\caption{\label{tab:g35_error_1} Other types of false negative errors made by GPT-3.5 AFaCTA on PoliClaim$_{test}$ other than not-enough-detail/context.}
\end{table*}

\begin{table*}[t]
\small
\centering
\resizebox{\textwidth}{!}{
\begin{tabular}{>{\raggedright\arraybackslash}m{0.13\textwidth}m{0.87\textwidth}}
\hline
\textbf{Error Type} & \multicolumn{1}{c}{\textbf{GPT-4 Errors and Explanations}} \\ \hline

False Positive: over-sensitive to granular, unspecific, or not-explicitly-presented facts & \textit{\textbf{Error 1:}} \hl{Requesting to work from home because of the \#coronavirus is what's called a ``reasonable accommodation.'' You have disabled people to thank for that. Remember this moment in history the next time you think Accessibility laws are too ``burdensome'' to be abided.} \textit{Error reason: the model recognizes the concept of ``reasonable accommondation'' and the existence of ``accessibility laws'' as facts, which are not explicitly presented by the post.}
 \\ \hline
 
False Negative: misunderstand verifiable fact as subjective interpretation & \textit{\textbf{Error 2:}} \hl{``Last week Trump told aides he's afraid journalists will try to purposefully contract \#coronavirus to give it to him on Air Force One.'' https://t.co/sS1MZR6D7w} \textit{Error reason: GPT-4 understands it as the tweet author's subjective interpretation of Trump's words. However, we think that it can be verified by checking whether Trump said the words or not.}

\textit{\textbf{Error 3:}} \hl{Due to \#coronavirus, media advises the economy must tank, the people must panic, Trump must be blamed, Biden must be secreted away from the public, and Bernie must cease rallies. I wonder why people do not trust the media's motives on this?} \textit{Error reason: GPT-4 understands it as the tweet author's subjective interpretation of the media's advice. However, we think it can be verified by checking if there are media suggesting such information.}
\\ \hline

\end{tabular}
}
\caption{\label{tab:t_g4_error} All errors made by GPT-4 AFaCTA on CheckThat!-2021-dev.}
\end{table*}

\begin{table*}[t]
\small
\centering
\resizebox{\textwidth}{!}{
\begin{tabular}{>{\raggedright\arraybackslash}m{0.13\textwidth}m{0.87\textwidth}}
\hline
\textbf{Error Type} & \multicolumn{1}{c}{\textbf{GPT-3.5 Errors and Explanations}} \\ \hline

False Negative: fail to identify facts entangled with opinions & \textit{\textbf{Error 1:}} \hl{Who would you prefer to lead our nation's response to the growing \#coronavirus threat?} \textit{Error reason: fail to identify ``the growing coronavirus threat''.}

\textit{\textbf{Error 2:}} \hl{It was a really really really really really really really really really really really really really really really really really really really really bad idea to elect Donald Trump President of the United States. \#TrumpVirus \#TrumpCrash \#TrumpRecession \#COVID19 \#coronavirus} \textit{Error reason: fail to identify ``elected Donald Trump Predisent of the US''.}

\textit{\textbf{Error 3:}} \hl{If people who are infected by corona virus in SA were black, their names, homes street: pictures will be all over Social Media. White privileges goes a long way. Wait for a case of a black person, they will mention even his location they won't say WC, they'll say Gugulethu ext 5} \textit{Error reason: fail to identify the correlation between the infected persons' race and their suffers.}

\textit{\textbf{Error 4:}} \hl{@realDonaldTrump On a morning when Americans are terrified, the markets are gonna historically crash and we need LEADERSHIP...all you've done is hate-tweet BULLSHIT about Sanders, Warren, Biden, Democrats, Schumer. the media and now Obama. Your incompetence is staggering.... \#Trump \#coronavirus} \textit{Error reason: although have subjective interpretations, the facts that ``markets are gonna historically crash'' and Trump commented something about others is verifiable.}

\textit{\textbf{Error 5:}} \hl{Dear BBC, I want to fight for you. I know you're more than news (which has been questionable) you're also great drama, documentaries, kids tv etc But don't make me question that by inviting Farage on to talk about Corona FFS!! Show you'll fight for your integrity yourselves!} \textit{Error reason: fail to identify the verifiable fact that BBC invited Farage is verifiable.}

\textit{\textbf{Error 6:}} \hl{Thread 1: One wonders about the racial politics of this corona outbreak. What would have happened had it been blacks who came into the country with the virus? Would they hav been allowed to ``self quarantine''? If the virus was from the continent; wouldn't travels be banned by now?} \textit{Error reason: fail to identify ``this corona outbreak'' and the practice of ``self-quarantine''.}

\textit{\textbf{Error 7:}} \hl{The total Iranian \#COVID19 case-count is in the hundreds of thousands, perhaps millions, according to my estimates (detailed at the link). This raises an important question: if there are two million cases, where are all the bodies? https://t.co/nHYbQlXlVC} \textit{Error reason: GPT-3.5 understands it as the tweet author's subjective interpretation about the numbers. However, it can be verified by checking details in the link and reliable data source.}

\textit{\textbf{Error 8:}} \hl{Due to \#coronavirus, media advises the economy must tank, the people must panic, Trump must be blamed, Biden must be secreted away from the public, and Bernie must cease rallies. I wonder why people don't trust the media's motives on this?} \textit{Error reason: GPT-3.5 understands it as the tweet author's subjective interpretation about the media's advice. However, we think it can be verified by checking if there are medias suggesting such information.}
\\ \hline

False Negative: fail to comprehend claims about attached links & \textit{\textbf{Error 9:}} \hl{Public Safety Announcement Fighting \#CoronaVirus. We have to do this together. Wishing good health to all of you! Love, Vijay. https://t.co/fbafmmtq8S} \textit{Error reason: this tweet claims that the link contains a public safetly announcement fighting COVID, which is verifiable.}

\textit{\textbf{Error 10:}} \hl{This thread needs to fly. It shows how the legacy media is USING covid-19 as a political weapon and even how the SAME reporters are contradicting themselves. This. Is. SICK. https://t.co/Werq544xii} \textit{Error reason: this tweet claims the content of the link, which is verifiable.}
\\ \hline

False Negative: fail to identify personal experience or citation & \textit{\textbf{Error 11:}} \hl{Beware the spread of coronavirus and Fox News pandemic propaganda. Trish Regan's melodramatic rant decrying Dems and MSM for allegedly exploiting \#COVID19 as ``another attempt to impeach the President.'' Yank this dangerous shrew off the air. \#Trumpdemic https://t.co/6B60RLMIS0} \textit{Error reason: fail to identify the personal experience of Trish Regan.}

\textit{\textbf{Error 12:}} \hl{I keep bumping into this problem. I want to be able to stand up and unequivocally defend the BBC. But it has repeatedly shut out radical voices and crucial issues while providing a massive platform for the alt-right to spout ill-informed nonsense. It is hard to love. https://t.co/A1pQMsqDxV} \textit{Error reason: The interpretation about BBC's behavior is subjective. But the tweet author's previous stance might be verifiable by checking their previous statements.}

\textit{\textbf{Error 13:}} \hl{@keywilliamss One African man actually got Corona and was cured in like a week. Health care officials are ``baffled as to why Africa is virtually unscathed.'' Which is... kinda racist that they expected it to be but lemme hush URL: https://t.co/yvP0DUXDiX} \textit{Error reason: fail to identify the African man's experience and the quotation of health care officials' speech.}

\\ \hline

\end{tabular}
}
\caption{\label{tab:t_g35_error} All errors made by GPT-3.5 AFaCTA on CheckThat!-2021-dev.}
\end{table*}

We conduct a thorough analysis on GPT-4 and GPT-3.5 AFaCTA. Errors on PoliClaim$_{test}$ can be found in \cref{tab:g4_error}, \cref{tab:g35_error}, and \cref{tab:g35_error_1}. Errors on CheckThat!-2021-dev can be found in \cref{tab:t_g4_error} and \cref{tab:t_g35_error}.

In both domains, we observe that GPT-4 is good at disentangling factual information from speeches or tweets. But it also leads to false positive errors due to over-sensitivity towards factual information. It also makes negative errors due to the lack of full context of the statements. In general, GPT-4 only makes mistakes on confusing samples that lie between factual and non-factual claims. 

GPT-3.5's errors concentrate on false negatives. It regularly hallucinates about personal experience and quotations which are explicitly defined in the prompts. It is very conservative in identifying anything as verifiable fact arguing there not enough ``specific details'' to determine verifiability. However, many facts are already specific enough for verification (see row 2 of \cref{tab:g35_error}). Sometimes, it also fails to identify facts entangled with opinions (see row 1 of \cref{tab:g35_error_1} and row 1 of \cref{tab:t_g35_error}). 

\end{document}